\documentclass[pdflatex,sn-mathphys-num]{sn-jnl}
\usepackage{graphicx}%
\usepackage{multirow}%
\usepackage{amsmath,amssymb,amsfonts}%
\usepackage{amsthm}%
\usepackage[mathscr]{eucal} 
\usepackage[title]{appendix}%
\usepackage{xcolor}%
\usepackage{textcomp}%
\usepackage{booktabs}%
\usepackage{mathtools}
\usepackage{tabularx, makecell, booktabs}
\usepackage{amssymb}%
\usepackage{pifont}
\usepackage{float}
\usepackage{adjustbox}
\usepackage{pgfplots} 
\usepgfplotslibrary{groupplots}
\usepackage[utf8]{inputenc}

\pgfplotsset{compat=1.18}

\newcolumntype{C}{>{\centering\arraybackslash}X} 
\newcommand{\cmark}{\ding{51}}%
\newcommand{\xmark}{\ding{55}}%

\newcommand{\R}{\mathbb{R}}
\newcommand{\E}{\mathbb{E}}
\newcommand{\bfu}{\mathbf{u}}

\newcommand{\bfc}{\mathbf{c}}
\newcommand{\bfy}{\mathbf{y}}
\newcommand{\bfb}{\mathbf{b}}
\newcommand{\bfK}{\mathbf{K}}

\newcommand{\bftheta}{\boldsymbol{\theta}}
\newcommand{\calD}{\mathcal{D}}

\makeatletter 
\def\@acknow{}%
\long\def\EarlyAcknow#1\par{%
\def\@acknow{\abstractfont\abstracthead*{Acknowledgments}%
#1\par}}%
\def\printabstract{\ifx\@acknow\empty\else\@acknow\fi\par%
    \ifx\@abstract\empty\else\@abstract\fi\par}
\makeatother

\definecolor{mycolor1}{rgb}{0.56, 0.0, 1.0}%
\definecolor{mycolor2}{rgb}{1.0, 0.75, 0.0}
\definecolor{mycolor3}{rgb}{0.0, 0.0, 1.0}%
\definecolor{mycolor4}{rgb}{0,1,0}%
\definecolor{mycolor5}{rgb}{1.0, 0.03, 0.0}%

\theoremstyle{thmstyleone}%
\newtheorem{theorem}{Theorem}%

\theoremstyle{thmstyletwo}%

\theoremstyle{thmstylethree}%

\begin{document}

\title[Article Title]{Weight-Parameterization in Continuous Time Deep Neural Networks for Surrogate Modeling}

\author*[1]{\fnm{Haley} \sur{Rosso}}\email{haley.rosso@emory.edu, \href{https://orcid.org/0009-0005-8583-4876}{0009-0005-8583-4876}}

\author[1]{\fnm{Lars} \sur{Ruthotto}}\email{lruthotto@emory.edu, \href{https://orcid.org/0000-0003-0803-3299}{0000-0003-0803-3299}}

\author[2]{\fnm{Khachik} \sur{Sargsyan}}\email{ksargsy@sandia.gov, \href{https://orcid.org/0000-0002-1037-786X}{0000-0002-1037-786X} }

\affil[1]{\orgdiv{Department of Mathematics}, \orgname{Emory University}, \orgaddress{\street{400 Dowman Dr}, \city{Atlanta}, \postcode{30307}, \state{GA}, \country{United States}}}

\affil[2]{\orgname{Sandia National Labs}, \orgaddress{\street{7011 East Ave}, \city{Livermore}, \postcode{94550}, \state{CA}, \country{United States}}}

\EarlyAcknow{
The authors would like to thank Sandia National Laboratories for funding this research and Emory University for supporting this work through computational resources and faculty mentorship. 
The authors would also like to thank Dr.\ Rebekah White for her valuable assistance in organizing and editing the manuscript.
}

\abstract{
Continuous-time deep learning models, such as neural ordinary differential equations (ODEs), offer a promising framework for surrogate modeling of complex physical systems. 
A central challenge in training these models lies in learning expressive yet stable time-varying weights, particularly under computational constraints. 
This work investigates weight parameterization strategies that constrain the temporal evolution of weights to a low-dimensional subspace spanned by polynomial basis functions.
We evaluate both monomial and Legendre polynomial bases within neural ODE and residual network (ResNet) architectures under discretize-then-optimize and optimize-then-discretize training paradigms. 
Experimental results across three high-dimensional benchmark problems show that Legendre parameterizations yield more stable training dynamics, reduce computational cost, and achieve accuracy comparable to or better than both monomial parameterizations and unconstrained weight models. 
These findings elucidate the role of basis choice in time-dependent weight parameterization and demonstrate that using orthogonal polynomial bases offers a favorable tradeoff between model expressivity and training efficiency.
}


\maketitle

\section{Introduction}\label{sec1}
The development of efficient surrogate models is essential for accelerating simulations of high-dimensional, computationally expensive systems, particularly those governed by ordinary differential equations (ODEs). 
Surrogate modeling aims to replace complex physical or numerical models with cheaper approximations that preserve essential input-output behavior. 
This is particularly important for tasks where thousands of model evaluations may be required, such as uncertainty quantification~\cite{Sudret2019} and data assimilation~\cite{KovachkiTgCNN2021}.

In recent years, deep neural networks (DNNs) have gained prominence as expressive and efficient surrogates in scientific machine learning contexts, especially for nonlinear systems where traditional reduced-order models may struggle~\cite{pinn_review, physicsinformeddeep, operatorlearning}.
Among these, continuous-time DNNs, such as neural ordinary differential equations (neural ODEs) and the continuous limit of residual networks (ResNets), offer promising surrogate modeling architectures. 
By modeling the forward pass of a neural network as the solution of an ODE, these architectures enable temporal smoothness and provide a structured framework for incorporating physical constraints or known system dynamics, improving interpretability and alignment with real-world processes.

In comparison to standard deep networks, continuous-time DNNs are particularly well-suited for modeling dynamical systems, as the network's evolution over time is governed by an explicit differential equation, allowing one to trace how inputs evolve through the network over time.
For surrogate modeling of ODE systems, prior information about the underlying physics can be incorporated into the network architecture, such as symmetries or conservation laws, which can lead to more accurate models~\cite{pinn_review, physicsinformeddeep}.
Continuous-time DNNs also provide a natural lens through which to apply techniques from control theory and numerical analysis. 
As a result, they are increasingly used to approximate dynamical systems or to emulate the solution operators of high-dimensional PDEs~\cite{ChenEtAl2018, ruthotto2024differential,gunther}.

While both neural ODEs and the continuous limit of ResNets make use of the same continuous framework, they are not equivalent. 
The first works specifically studying the continuous limit of ResNets, to our knowledge, are~\cite{E:2017kz, HaberRuthotto2017}; in~\cite{HaberRuthotto2017}, the learning problem is given as an optimization problem where the weights are updated using the Gauss-Newton preconditioned conjugate gradient (PCG) method.
The continuous limit of a ResNet, often derived from interpreting discrete residual layers as time steps, allows the weights to vary across layers and, by extension, over time~\cite{ruthotto2024differential,HaberRuthotto2017}. 
On the other hand, neural ODEs typically assume that the network dynamics are governed by a differential equation with time-invariant weights, which are optimized through an adjoint-based continuous-time training procedure~\cite{ChenEtAl2018}. 
This key difference affects both model expressivity and numerical behavior.

For instance, discretizing a neural ODE using an explicit scheme such as forward Euler or Runge-Kutta may yield poor approximations, especially with large step sizes, since its static-weight formulation cannot capture the same layer-to-layer variability as a time-varying ResNet~\cite{ott2021resnet,sander2022residual}. 
These structural differences also influence training pipelines; ResNets are often discretized first and then optimized, whereas neural ODEs are typically optimized in continuous time and then discretized for inference. 
Our work addresses both approaches using a unified framework based on time-dependent weight parameterization.

In any case, the benefits offered by continuous-time DNNs come with a cost.
Training continuous-time deep neural networks such as neural ODEs or deep ResNets often involves optimizing high-dimensional parameter spaces, particularly when time-varying weights are discretized at many points or parameterized with expressive basis functions.
%
%
These difficulties are exacerbated in surrogate modeling for PDE-governed systems, which often involve stiff dynamics themselves, such as in convection–diffusion–reaction models~\cite{gunther, ChenEtAl2018, Sudret2019}.

To mitigate these challenges, one promising strategy is to parameterize the time-dependent weights of the network using a small, pre-determined number of basis functions. 
Such weight parameterization can be interpreted as a form of model reduction, enforcing smoothness in the network while reducing the dimensionality of the parameter space and improving generalization. 
For example, B-spline parameterizations have been used to regularize training and promote stability~\cite{gunther}, while polynomial basis functions such as Legendre or Chebyshev polynomials offer orthogonality properties that support efficient optimization~\cite{autonomous}.

Despite these advantages, the impact of weight parameterization on surrogate modeling performance --- particularly across different continuous-time architectures and polynomial basis functions --- remains underexplored. 
In addition, most studies either discretize the network first and then optimize (common in ResNets), or solve an optimal control problem in continuous time (typical in neural ODEs)~\cite{Li:2019wr, benning2019deep, ruthotto2024differential}, but few compare these approaches side-by-side under different parameterization regimes.
Moreover, many works do not explicitly treat time as a network input or explore the use of orthogonal basis functions for temporal weights, especially in high-dimensional, PDE-based surrogate modeling tasks.

In this work, we address these gaps by studying polynomial weight parameterization for continuous-time deep neural networks in the context of surrogate modeling for PDE systems. 
We explore two families of basis functions, monomials and Legendre polynomials, and apply them to both ResNets and neural ODEs. 
In our test cases, we also include a Hamiltonian-inspired system with forward propagation resembling a symmetric, second-order ODE modeled after the continuous version of a ResNet layer in~\cite{HaberRuthotto2017}. This system bears resemblance to a ResNet, with weights that vary with each layer, but utilizes a Verlet time integrator.

To evaluate the efficacy of our proposed approaches, we considered surrogate modeling tasks tied to real-world scenarios such as climate modeling and physical phenomena.
Specifically, we utilized the popular, high-dimensional training data sets ELM (E3SM Land Model), a component of the DOE Energy Exascale Earth System Model (E3SM) project, alongside convection diffusion reaction (CDR) and direct current resistivity (DCR).
The CDR is a system of PDEs which measure various physical phenomena, and the DCR data represents an inverse conductivity problem using a PDE that models electric potential~\cite{VarPro}.

Our key contributions are as follows:
\begin{itemize}
    \item We develop parameterized formulations for both ResNets and neural ODEs using monomial and Legendre basis expansions of time-varying weights, comparing discretize-then-optimize and optimize-then-discretize training approaches.
    \item We evaluate the expressivity, accuracy, smoothness, and computational cost of these architectures on three high-dimensional surrogate modeling problems: the E3SM Land Model (ELM), a convection-diffusion-reaction (CDR) system, and a direct current resistivity (DCR) inverse problem.
    \item We provide a quantitative analysis of the trade-offs associated with different weight parameterizations, highlighting scenarios where low-dimensional representations reduce training cost and improve stability without sacrificing accuracy.
\end{itemize}

Through this, our work contributes new insights into how basis function choice and parameterization strategy affect performance in continuous-time neural networks used for surrogate modeling. 
In particular, we demonstrate that Legendre polynomial parameterization can reduce the number of trainable weights while maintaining expressivity, particularly in the neural ODE setting where the number of function evaluations dominates runtime.
By targeting real-world datasets, this study bridges theoretical developments in weight parameterization with practical surrogate modeling challenges.




\subsection{Outline}
The outline of this paper is as follows. 
Section~\ref{sect:lit} describes the existing literature on the topic of weight parameterization for neural ODE and ResNets.
In Section~\ref{sect:TrainingAlgs}, we outline the differences between the optimize-then-discretize and discretize-then-optimize approaches, and we define the continuous learning problem.
We establish the weight parameterization methods in Section~\ref{sect:WeightParam}. 
Section~\ref{sect:results} includes results of our computational experiments, followed by a discussion of these results in section~\ref{sect:discussion}.
Finally, we conclude in section~\ref{sect:future} and suggest future directions for this work.
Corresponding tables and figures can be found in sections~\ref{sect:tables} and~\ref{sect:Figures}, respectively.

\section{Related Literature}\label{sect:lit}

Continuous-time deep neural networks, such as neural ordinary differential equations (neural ODEs) and the continuous limit of ResNets, have gained attention in recent years as continuous representations of network evolution, particularly due to their ability to approximate dynamical systems governed by ODEs or PDEs.
This is especially useful in surrogate modeling applications, where training data may be limited, temporal smoothness is important, and high-dimensional systems require efficient representations~\cite{ruthotto2024differential, gunther,zhong2019symplectic}.

Unlike standard discrete-layer networks, continuous-time models offer a way to preserve the underlying dynamics and reduce the need for large parameter sets, making them attractive candidates for surrogate modeling of high-dimensional systems; several recent works showcase this potential.
In~\cite{nair2024investigation}, the authors use an autoencoder and neural ODE framework to model complex PDE systems (e.g., Kuramoto-Sivashinsky, compressible Navier-Stokes). 
The autoencoder reduces dimensionality, and dynamics are learned in latent space with a neural ODE.\@
They find that the latent neural ODE successfully captures key dynamical timescales of the full system and provides accelerated surrogate evaluations while maintaining accuracy.
Moreover, analysis of the Jacobian eigenvalues reveals how training trajectory length influences the model's performance.

The work~\cite{Vermari_n_2025} introduces a latent augmented neural ODE surrogate designed to emulate the costly chemical reaction component in a 3D molecular cloud simulation (3D-PDR).
They find that the neural ODE surrogate replicates key outputs like column density maps accurately and runs significantly faster than the original chemical solver, making it practical for large-scale simulations.
The paper~\cite{Zhou_2025_neuralPDE} proposes a neural surrogate architecture that predicts temporal derivatives, rather than next states, combined with classical ODE integration for time stepping.
This design increases stability and allows for flexible time stepping during inference and outperforms traditional black-box surrogates in accuracy and robustness.

These works demonstrate that continuous-time DNNs can serve as effective surrogate models for high-dimensional ODE systems.
To better understand how these surrogate models are constructed and trained, we now turn to the broader neural ODE literature, which can be categorized based on how the model handles weight dynamics.

More general literature on neural ODEs can be broadly categorized into two classes based on how the weights in the ODE function are treated: static weights~\cite{ChenEtAl2018, ott2021resnet}, and time-dependent weights~\cite{gunther, generalized, autonomous, dissect}.
This distinction is essential because it directly influences model expressivity, training behavior, and suitability for surrogate modeling tasks. 
Below, we outline how these approaches are formulated and applied, and how our work builds upon and extends this foundation in the context of surrogate modeling.

The most widely cited formulation of neural ODEs is presented in~\cite{ChenEtAl2018}, where the weights of the neural network are static, and time-dependence in the learned dynamics is introduced by explicitly including time as an input feature. 
This formulation can be viewed as a continuous analogue of ResNets, and has inspired a wide range of work in the area.
However, as noted in~\cite{dissect, autonomous}, the static-weight formulation found in~\cite{ChenEtAl2018} can lead to a mismatch between the continuous neural ODE and its intended discrete ResNet counterpart unless specific architectural conditions are met~\cite{ott2021resnet, sander2022residual, heEtAl}. 
In other words, static-weight neural ODEs may not reduce to standard residual networks when discretized, making their interpretability and implementation less seamless in some settings. 
This has led to increased interest in time-dependent weight formulations, which are more naturally aligned with residual architectures.

An example of time-dependent weights is found in the work of~\cite{HaberRuthotto2017}, where weights are treated as piecewise functions of time. 
In this discretize-then-optimize approach, the network is discretized first (e.g., via forward Euler), and weights are learned at each time step as part of the optimization problem. 
This strategy is attractive for surrogate modeling because it naturally recovers ResNets when using unit step sizes and simple integration schemes.
However, the number of parameters in this setup scales with the number of time steps, which can rapidly become computationally expensive, especially in high-dimensional systems.
To mitigate this, regularization is commonly imposed on the weights over time to enforce stability and smoothness~\cite{HaberRuthotto2017, dissect}.

To reduce parameter count while maintaining time-dependence and stability, recent works have introduced explicit parameterizations of the weights as functions of time. 
These parameterizations serve two main purposes: to encode smooth temporal dynamics directly into the model and to reduce the dimension of the trainable parameter space.
This approach is particularly useful for optimize-then-discretize methods, where one first optimizes a continuous neural network
$f(t,x;\bftheta(t))$ and then discretizes the resulting ODE.\@
However, as we demonstrate in this work, parameterized weights can also yield substantial benefits in discretize-then-optimize frameworks, especially in surrogate modeling tasks that demand a balance between expressivity and efficiency.

Weight parameterization approaches can be grouped by their parametric model. 
The models we primarily focus on are polynomials, such as in~\cite{autonomous}, but other parameterization options such as neural networks and splines are presented in works such as in~\cite{gunther,generalized}.
In~\cite{autonomous}, the authors propose NANODE, a neural ODE model where weights are expanded in polynomial bases such as monomials, Chebyshev, and Legendre polynomials. 
This allows for expressive modeling with interpretable, structured variation in time.
In~\cite{dissect}, a similar polynomial basis approach is used for parameterization, but notably the weights in that work are not time-dependent—meaning they vary spatially (in network depth) but not as a function of continuous time.
In~\cite{gunther}, time-dependent weights are modeled using B-spline basis functions. 
This decouples the parameters from individual layers and instead defines them globally over time, enhancing both smoothness and generalization. 
The reduced parameter count also improves efficiency, and the approach has been found to increase training stability.

Another theme that intersects with time-dependent weight parameterization is the introduction of orthogonality constraints on the weights. 
Several studies have shown that enforcing orthogonality helps preserve gradient norms and mitigates the vanishing/exploding gradient problem, especially in deep and recurrent architectures~\cite{Huang_Liu_Lang_Yu_Wang_Li_2018, orthogonalityrecurrentnetworks, autonomous}.
~\cite{Huang_Liu_Lang_Yu_Wang_Li_2018} show that orthogonal weights boost performance in ResNets on standard image datasets.
~\cite{orthogonalityrecurrentnetworks} explores the trade-offs of orthogonality: while it aids stability, strict constraints can reduce expressiveness and slow convergence.
Moreover,~\cite{autonomous} finds that orthogonalization significantly improves training stability when using Chebyshev polynomial bases to parameterize time-varying weights.

Despite the promising directions above, there are limitations that remain unaddressed.
For instance, while~\cite{gunther, HaberRuthotto2017} incorporate time-dependent weights, they do not model time as an input to the function.
This limits the model's ability to learn dynamics that explicitly depend on time, rather than just evolving implicitly over layers or steps.
Furthermore, although polynomial bases have been used for parameterization, there is limited comparative analysis across different basis types (e.g., monomial vs.\ orthogonal polynomials) and little investigation into their performance in high-dimensional surrogate modeling tasks.
Additionally, most works adopt either discretize-then-optimize or optimize-then-discretize paradigms exclusively. 
As such, there remains room to explore these paradigms and their applicability to surrogate modeling tasks.

Our work addresses this by incorporating time as an explicit input in the neural network $f$ (section~\ref{sect:continuousproblem}), exploring both time-dependent and time-independent weight parameterizations using multiple basis functions, including monomials and Legendre polynomials (section~\ref{sect:WeightParam}), and applying our approach to three high-dimensional surrogate modeling problems (section~\ref{sect:results}), comparing performance across parameterizations and training algorithms, and evaluating both standard ResNets and Hamiltonian-inspired architectures.
Our contributions in comparison to other works are summarized in table~\ref{tab:literature}.
This comprehensive study sheds light on the trade-offs between expressivity, stability, and computational cost, while offering new insights into the benefits of basis parameterization in neural ODEs in continuous-time surrogate modeling.

\section{Numerical Techniques for Training}\label{sect:TrainingAlgs}
As mentioned in section~\ref{sect:lit}, the distinction between weights that are time-dependent or static influences the choice between two fundamental training approaches, delineated by~\cite{discoptvsoptdisc}:
\begin{itemize}
    \item \textbf{Discretize-then-optimize}: the weights are first discretized in time, transforming the problem into a finite-dimensional optimization problem that is subsequently solved.
    \item \textbf{Optimize-then-discretize}: the objective function is minimized in the continuous setting before discretization is applied to approximate the learned weights.
\end{itemize}
In this section, we will first define the continuous learning problem that is common to both approaches, and then we will discuss the numerical methods used to solve the problem in sections~\ref{sect:optdisc} and~\ref{sect:discopt}.

In general, we consider a neural ODE or ResNet as a function $F:\R^n \to \R^m$ that maps input data $\bfy\in\R^n$ to output data $\bfc\in\R^m$.
The design of the nonlinear function $f$ from~\eqref{eq:IVP} dictates the model dynamic and significantly impacts training outcomes.
The most suitable choice of training method depends on both the how the weights are parameterized and the optimization strategy; the following sections provide a more detailed analysis of these choices.
Regardless of the approach, the end goal is to optimize the network weights such that the learned model best approximates a high-fidelity surrogate model (surrogates are described in section~\ref{sect:results}).

\subsection{The Continuous Learning Problem}\label{sect:continuousproblem}
To define the learning problem that relates to both approaches (optimize-then-discretize and discretize-then-optimize), we first consider a training set of samples from some distribution $\calD$ of labeled pairs $(\bfy,\bfc) \in \R^n \times \R^m$.
Our goal is to train the weights $\bftheta$ of our neural network $F:\R^n \to \R^m$ such that $F(\bfy,\bftheta) \approx \bfc$. 

In the context of continuous dynamics, as presented by Chen et al.~\cite{ChenEtAl2018}, the evolution of the hidden layers of $F$ can be represented by a nonlinear and nonautonomous ``neural ODE'' that solves the initial value problem (IVP):
\begin{equation}\label{eq:IVP}
    \frac{d}{dt} \bfu(t) = f(\bfu(t), t, \bftheta_{\rm NODE}(t)), \quad t\in(0,T], \quad \bfu(0) = \sigma(\bfK_{\rm in} \bfy + \bfb_{\rm in}).
\end{equation}
where $\bfu(t)$ denotes the state of the network, and the activation function $\sigma$ defines the initial conditions. 
Compared to~\cite{ChenEtAl2018}, however, we generalize this model to allow the weights $\bftheta_{\rm NODE}(t)$ to be time-dependent, allowing us to structure the training process as an optimal control/variational problem as in~\cite{gunther}, given by
\begin{equation}\label{eq:control_problem}
\begin{split}
   \min_{\bftheta}\ell(\bftheta) \coloneq \E_{(\bfy,\bfc)\sim\mathcal{D}}\left[ \frac12 \| F(\bfy,\bftheta) - \bfc \|^2\right] + \frac{\alpha}{2}\|\bftheta\|^2 \\
   \text{subject to} \quad \frac{d}{dt} \bfu(t) = f(\bfu(t), t, \bftheta_{\rm NODE}(t)), \quad t\in(0,T], \\ 
   \bfu(0) = \sigma(\bfK_{\rm in} \bfy + \bfb_{\rm in}).
\end{split}
\end{equation}
Here, $\bftheta$ represents the weight matrix $\bfK_{in}$, the bias vector is $\bfb_{in}\in\R^n$, and the time-dependent weights $\bftheta_{\rm NODE}$.
The regularization term with coefficient $\alpha$ prevents overfitting by penalizing large weights.

The time-dependent weights $\bftheta_{\rm NODE}$ define the control function that is learned during the optimization process.
The dimension of $\bftheta_{\rm NODE}$ may be infinite, depending on the problem of interest.
The objective is to minimize the expected error between the output of $F$ and the target output data across the entire training distribution, subject to the dynamics of the neural network as modeled by the ODE.\@

\subsection{Optimize-then-discretize}\label{sect:optdisc}

%
For the optimize-then-discretize method, one first optimizes the time-dependent weights $\bftheta_{\rm NODE}(t)$ in the continuous setting, and then discretizes the ODE~\eqref{eq:IVP} to obtain a finite-dimensional approximation of the neural network.
This requires solving the forward ODE~\eqref{eq:IVP} for $\bfu(t)$, which is then used to compute the loss functional $\ell(\bftheta)$ in~\eqref{eq:control_problem}.
The loss functional is minimized using gradient descent, where the gradients are computed backward in time  using the adjoint method.

The adjoint $\mathbf{a}(t)$ is defined as the gradient of the loss functional $\ell(\bftheta)$ with respect to the state $\bfu(t)$, which is given by
\begin{align*}
        \mathbf{a}(t) &= \frac{\partial \ell}{\partial \bfu(t)}
\end{align*}
The adjoint $\mathbf{a}(t)$ evolves backward in time, starting from the final time $T$ and moving to the initial time $0$.
It satisfies the continuous adjoint equation, which is derived from the chain rule and the definition of the loss functional:
\begin{align*}
        \dot{\mathbf{a}}(t) &= -\left( \nabla_{\bfu} f{\left(\bfu(t), t, \bftheta_{\rm NODE}(t)\right)}^{T}\right)\mathbf{a}(t) 
\end{align*}
where $\nabla_\bfu f \in \mathbb{R}^{n\times n}$ is the Jacobian of $f$ with respect to the state $\bfu(t)$. 
The transpose ${(\nabla_\bfu f)}^{T}$ ensures dimensional consistency when applying the chain rule in reverse-mode differentiation. 
This formulation arises naturally from the calculus of variations or the Pontryagin Maximum Principle in optimal control~\cite{Pontryagin_Max}.


The gradient of loss with respect to the parameters $\bftheta$ can be computed using the adjoint method, which allows us to backpropagate through the ODE solution $\bfu(t)$.
Combining the mean-squared error term with the regularization gradient $\nabla_{\bftheta} \ell_{\text{REG}} = \alpha\bftheta$, we obtain the gradient of the loss with respect to $\bftheta$,
\begin{equation*}
    \nabla_{\bftheta} \ell(\bftheta) = \int_0^T {\textbf{a}(t)}^T\frac{\nabla_{\bftheta} f(\bfu(t), t, \bftheta_{\rm NODE}(t))}{\mathbf{a}(t)}\,dt + \alpha\bftheta.
\end{equation*}
A more detailed derivation in the context of neural ODEs can be found in~\cite{ChenEtAl2018}.

The state $\bfu(t)$ is first computed forward in time using an adaptive time integrator, which allows us to obtain estimates of $\bfu$ at specific time points.
Here, it is crucial to obtain accurate estimates of $\bfu(t)$, since implicit differentiation will only work on points that lie exactly on the manifold, i.e., the set of all paths that satisfy the ODE~\eqref{eq:IVP}.
After this forward pass, we compute $\mathbf{a}(T)$ to initiate the adjoint method backward in time.
This allows us to compute the gradients of the loss functional with respect to the state $\bfu(t)$, which can then be used to compute the gradients with respect to the time-dependent weights $\bftheta_{\rm NODE}(t)$.
These gradients are expressed as continuous integrals over the time interval $[0,T]$.

Since closed-form solutions for $\bfu(t)$ are not available, we introduce numerical methods to discretize both the forward and adjoint ODEs. 
Up to this point, $\bfu$ has been optimized in a continuous setting, but we now need to discretize the ODE to obtain a finite-dimensional approximation of the neural network.
The numerical method we use in this work is the Dormand-Prince method for time integration, a blend of an order 4 and order 5 explicit Runge-Kutta method. More details can be found in~\ref{sect:results}.

One challenge of the adjoint method is that it can be prohibitively expensive, as it requires storing all ODE solutions across the entire time interval. 
Additionally,~\cite{BIROS_ANODE, HaberRuthotto2017} point out that neural ODEs are not always reversible, and backward integration can introduce numerical instability. 
Therefore, we must carefully consider stability issues during both the forward and backward passes.

\subsubsection{Stability}\label{subsubsect:stability_prob} 

Stiff ODE solvers are designed to maintain numerical stability over long time horizons, even in the presence of rapidly changing dynamics. 
Nonetheless, it is essential to consider stability enforcement in both the forward and backward computations, particularly when using neural ODEs, where solver choice and step size can influence gradients.
We begin by recalling the Picard-Linderlof theorem:

\begin{theorem}[Picard-Linderlof theorem] Let \( D \subseteq \mathbb{R} \times \mathbb{R}^{n} \) be a closed rectangle with  
\( (t_{0}, y_{0}) \in \operatorname{int} D \), the interior of \( D \).  
Let \( f: D \to \mathbb{R}^{n} \) be a function that is continuous in \( t \) and Lipschitz continuous in \( y \) (with a Lipschitz constant independent of \( t \)).  
Then, there exists some \( \varepsilon > 0 \) such that the initial value problem  

\[
y'(t) = f(t, y(t)), \qquad y(t_{0}) = y_{0}
\]
has a unique solution \( y(t) \) on the interval \( [t_{0} - \varepsilon, t_{0} + \varepsilon] \)~\cite{abraham_manifolds_1993}.
\end{theorem}
Since $f$ is assumed to be Lipschitz, this theorem guarantees local existence and uniqueness of the solution. 
Moreover, it implies that the flow of the ODE is locally reversible—meaning that, for sufficiently small time intervals, one can integrate backward to recover the initial state~\cite{calcaterra2006lipschitzflowboxtheorem}.
However, this reversibility generally fails to hold over long time horizons in practice due to the accumulation of numerical errors and the inherent instability of backward integration, particularly in high-dimensional, nonlinear systems.

As discussed in section~\ref{sect:optdisc}, training neural ODE models with continuous adjoint methods involves solving a backward-in-time problem for the adjoint state. 
This backward pass requires integrating an adjoint ODE that depends on both the original state trajectory and the Jacobian of the forward dynamics.
In the special case of linear, autonomous systems with $f(y)=Ay$ (a linear and autonomous ODE), the backward dynamics can be written as $\dot{y}=-Ay$, resulting in eigenvalues of the Jacobian $J_f$ being negated. 
In such settings, the stability of the forward dynamics corresponds to eigenvalues satisfying $\Re{\lambda}<0$, leading to decay over time~\cite{HaberRuthotto2017}. 
Reversing time flips the sign of these eigenvalues, potentially inducing exponential growth and instability in the backward pass~\cite{Khalil_2002}.

However, in general, particularly when $f=f(t,y)$ is nonlinear or time-dependent, the Jacobian $J_f(t,y)$ varies over time, and the eigenvalue-based reasoning becomes insufficient. 
Stability analysis in such settings requires more advanced tools, such as Lyapunov functions or examining the spectral properties of the monodromy matrix over an interval. 
Numerical instability may still arise during backward integration due to stiffness, chaotic sensitivity, or poor solver conditioning.

To mitigate these issues, various stabilization techniques may be employed. 
These include regularizing the learned vector field $f$, restricting the Jacobian spectrum through architectural constraints, or using solver-adaptive adjoint strategies such as checkpointing or discrete adjoints~\cite{BIROS_ANODE}.
Enforcing stability in both forward and backward computations is therefore essential for reliable and efficient training of neural ODEs.

One approach to ensuring stability is through the use of a reversible architecture, such as a Hamiltonian ODE.\@
This formulation, which resembles a symmetric second-order ODE, can be discretized using a symplectic integrator such as the Verlet method~\cite{HaberRuthotto2017}. 
Symplectic integrators preserve geometric structure and ensure stability in both the forward and backward passes. 
In contrast, explicit methods like forward Euler are unstable for stiff systems, and even higher-order methods like Runge-Kutta 4 exhibit limited stability, as they only cover a small region in the complex plane. 
The Verlet integrator, however, conserves the Hamiltonian, providing superior long-term stability compared to standard non-geometric integrators~\cite{ruthotto2020numerical}.
We present numerical results demonstrating the effectiveness of the Hamiltonian architecture in Section~\ref{sect:results}.

An alternative memory-efficient approach is the ANODE method introduced in~\cite{BIROS_ANODE}, which incorporates a ``checkpointing'' strategy to reduce memory complexity from $\mathcal{O}(Ln)$, where $L$ is the number of layers and $n$ is the number of time steps, to $\mathcal{O}(L) + \mathcal{O}(n)$. 
This guarantees backward stability while maintaining the same computational complexity as the adjoint-based method in~\cite{ChenEtAl2018}. 
Instead of storing the entire trajectory (which is memory-intensive) or relying purely on backward integration (which is unstable), ANODE saves specific intermediate states (``checkpoints'') during the forward pass.
Between checkpoints, the forward pass is recomputed in small segments, to help avoid exacerbating errors when integrating backward over long time horizons.
Because each segment of the ODE is recomputed forward in time (rather than extrapolating backward from the final state), the adjoint computation does not suffer from the same numerical instability that arises when reversing a stiff system.

The checkpointing method falls somewhere between discretize-then-optimize and optimize-then-discretize. 
ANODE does not fully follow a strict discretize-then-optimize approach because it does not store the full trajectory of discretized states and instead recomputes segments of the time series during the backward pass.
Moreover, it does not constitute a strict optimize-then-discretize approach because it uses checkpointing and local recomputation, introducing a hybrid discretization strategy.

\subsection{Discretize-then-optimize}\label{sect:discopt}

For a ResNet, we discretize the ODE~\eqref{eq:IVP} using the forward Euler method, parameterize $\bftheta_{\rm NODE}(t)$ in time, and approximate the network state 
$\bfu$ to transform the continuous problem into a finite-dimensional optimization problem.

In this context, the state $\bfu$ and control $\bftheta$ can be integrated using different numerical schemes. 
For example, as shown in~\cite{Li:2019wr, benning2019deep}, some methods parameterize ResNet weights with a spline, where the nodes of the spline differ from those of the state's. 
This is permissible because of our choice of $f$ in~\eqref{eq:IVP}, where the time-dependent bias within the nonlinearity can be treated separately from the weights as an input feature. 
Specifically, the bias is parameterized as a function of time, and the last column of the weight matrix (described in Section~\ref{sect:WeightParam}) corresponds to the bias vector.

The update rule for the forward Euler discretization of the ODE~\eqref{eq:IVP} is defined as
\begin{equation}{\label{eq:fwd_euler}}
    \bfu^{n+1} = \bfu^n + \Delta t \, f(\bfu^n, t_n, \bftheta_{\rm NODE}(t_n)), \quad n = 0, \dots, N-1, \quad     \bfu^0 = \sigma(\bfK_{\rm in} \bfy + \bfb_{\rm in}).
\end{equation}
Each network layer corresponds to a time step of the ODE, producing a sequence of discrete, layer-wise updates that transforms the continuous problem into a discrete optimization task over the parameters $\bftheta_{\rm NODE}$.
The discretized ODE~\eqref{eq:fwd_euler} can be viewed as a residual network, where the weights $\bftheta_{\rm NODE}$ are time-dependent and parameterized as described in section~\ref{sect:WeightParam}.

The goal of the optimization step is to compute the gradient of the loss with respect to the weights $\bftheta_{\rm NODE}$.
To achieve this, we use backpropagation, recursively computing the gradient from $ \bfu^N $ back to $\bfu^0 $, accumulating gradients at each step. 
%
Once we have the necessary gradients through backpropagation, we use ADAM to iteratively update the weights. 
ADAM, an adaptive moment estimation algorithm, computes adaptive learning rates by estimating the first and second moments of the gradients~\cite{adam}. 
This method smooths the gradient updates using momentum and adjusts step sizes based on the magnitudes of the gradients.

In addition to ADAM, we also consider the Gauss-Newton variable projection (GNvpro) method, an extension of variable projection to non-quadratic loss functions~\cite{VarPro}. 
GNvpro optimizes over a subset of parameters by exploiting the structure of least-squares problems, splitting parameters into linear and nonlinear components. 

The choice of time integrator and network architecture in the discretize-then-optimize setting impacts the stability of the problem.
For example, networks inspired by Hamiltonian mechanics described in section~\ref{subsubsect:stability_prob}, which have a Jacobian with imaginary eigenvalues, can become unstable when using a method such as forward Euler~\cite{HaberRuthotto2017}. 

\medskip

For both procedures outlined in sections~\ref{sect:optdisc} and~\ref{sect:discopt}, the common goal is estimating the network weights to approximate high-fidelity data in a way that is efficient and accurate. 
Thus, we introduce weight parameterization and investigate the impact of different parameterization types on the optimization and discretization processes.

\section{Weight-Parameterization}\label{sect:WeightParam}

In both neural ODEs and residual networks, the weight functions $ \bftheta(t) \in \mathbb{R}^n $ vary over time.
However, representing a distinct trainable weight at every time step is computationally expensive and may lead to overfitting or unstable training dynamics. 
To address this, we constrain the weights to lie in a low-dimensional function space spanned by a fixed set of basis functions. 
This reduces the number of trainable parameters while enforcing smoothness in time.

In section~\ref{sect:continuousproblem}, we define the weights of our problem as $\bftheta_{\rm NODE}(t)$, which are potentially infinite-dimensional.
These weights may introduce numerical challenges, such as instability, lack of smoothness, and high computational cost.
One way to mitigate these issues is to parameterize the weights as functions of time, 
which allows us to guarantee finite-dimensionality, helping to control their complexity and smoothness.
Parameterization also enables us to reduce the number of trainable parameters, 
making the optimization process more efficient and stable.

In this work, we enforce weight parameterization by expressing the time-dependent weights $\bftheta_{\rm NODE}(t)$ as a linear combination of a fixed set of polynomial basis functions, depicted in equation~\eqref{eq:weightParam}.
By being able to choose the polynomial degree $d$, we can control the number of trainable parameters, which is particularly useful for optimizing high-dimensional problems.
We study the role of weight parameterization in both the discretize-then-optimize and optimize-then-discretize approaches in the context of the supervised function approximation problem from section~\ref{sect:continuousproblem}. 

The time-parameterized weight vectors $\bftheta_{\rm NODE}(t) = \{\bftheta_1, \dots, \bftheta_N\} \in \mathbb{R}^n$ for the neural network $F(\bfy, \bftheta) = \bfy_N$ are expressed as the following sum:

\begin{equation}
\bftheta_P(t) = \sum_{i=1}^d p_i(t) \bftheta_i    
\label{eq:weightParam}
\end{equation}
where ${\{p_i(t)\}}_{i=1}^d$ is a fixed set of polynomial basis functions (e.g., monomial or Legendre), and $\bftheta_i \in \mathbb{R}^n$ are trainable coefficient vectors.
We select the polynomial degree $d$ and the number of time points $N$ for evaluation.
This formulation allows us to generate all time-varying weights from a small number of basis coefficients, improving both computational efficiency and generalization.

The choice of basis functions plays a critical role; while monomials can lead to numerical instability due to poor conditioning (see section~\ref{subsect:weight_param_stability}), orthogonal polynomials such as Legendre provide better-behaved interpolants with stable derivatives and reduced overfitting.
We apply this parameterization identically in both the discretize-then-optimize (ResNet-style) and optimize-then-discretize (neural ODE) training paradigms.

\subsection{Discretize-then-optimize --- weight parameterization}\label{subsect:discopt_weightParam} 
In this discretize-then-optimize context, weight parameterization is not explicitly necessary, as the weights are discretized at a fixed set of time points ${\{t_j\}}_{j=0}^{N-1}$.
However, it still offers benefits such as reduced parameter count and improved training stability.

To set up the problem with parameterized weights, we first define the basis functions $p_i(t)$, which are evaluated at the discrete set of $N$ time points ${\{t_j\}}_{j=0}^{N-1}$.
We then form a matrix $\textbf{A} \in \mathbb{R}^{d \times N}$, where each entry is the evaluation of the $i^\text{th}$ basis function at the $j^\text{th}$ time step:
\begin{equation*}
\textbf{A}_{ij} = p_i(t_j).
\end{equation*}
We then collect the coefficient vectors into a matrix $\mathbf{\Theta} \in \mathbb{R}^{n \times d}$, with each column corresponding to one trainable coefficient vector $\bftheta_i$. 
As a result, the original neural network $F(\bfy,\bftheta)$ becomes
\begin{equation}
F(\bfy, \mathbf{\Theta} \textbf{A}).
\label{eq:NewNetwork}  
\end{equation}

Once we have this parameterized weight representation, the network weights from~\eqref{eq:NewNetwork} can be written as a matrix product
\begin{equation}
\begin{bmatrix} {\bftheta_P(t_1)}^{(1)} & {\bftheta_P(t_2)}^{(1)} & \cdots & {\bftheta_P(t_N)}^{(1)}\\
\vdots & \ddots & \vdots \\
{\bftheta_P(t_1)}^{(n)} & {\bftheta_P(t_2)}^{(n)} & \cdots & {\bftheta_P(t_N)}^{(n)}\end{bmatrix} = \begin{bmatrix} \bftheta_1^{(1)} & \bftheta_2^{(1)}& \cdots & \bftheta_d^{(1)}\\
\vdots & \ddots & \vdots \\
\bftheta_1^{(n)} & \bftheta_2^{(n)}& \cdots & \bftheta_d^{(n)} \end{bmatrix} \begin{bmatrix}p_1(t_1)& \cdots & p_1(t_N) \\ p_2(t_1) & \cdots & p_2(t_N)\\ \vdots & \ddots & \vdots \\  p_d(t_1) & \cdots & p_d(t_N) \end{bmatrix}.
\label{eq:MatrixProduct}
\end{equation}
Here, the first matrix on the left-hand side contains the time-parameterized weights $\bftheta_P(t_j) \in \mathbb{R}^n$ for each time point $t_j$, and the second matrix on the right-hand side contains the trainable coefficients $\bftheta_i \in \mathbb{R}^n$ for each basis function $p_i(t)$.
The right-hand side matrix is the basis function matrix $\textbf{A}$, which contains the evaluations of the basis functions at each time point.

More succinctly, the full set of time-discretized weights utilized in~\eqref{eq:MatrixProduct} can be compactly rewritten as a matrix product $\mathbf{\Theta} \textbf{A}$ such that
\begin{equation}
\bftheta(t_j) = \mathbf{\Theta} \textbf{A}_{:,j}, \quad \text{or} \quad [\bftheta(t_0), \dots, \bftheta(t_{N-1})] = \mathbf{\Theta} \textbf{A}
\label{eq:theta_matrix_product}
\end{equation}
where $\mathbf{\Theta} \in \mathbb{R}^{n \times d}, \; \textbf{A} \in \mathbb{R}^{d \times N}$, and $\textbf{A}_{ij} = p_i(t_j)$.
When $\bftheta$ is multiplied by $\textbf{A}$ in~\eqref{eq:NewNetwork}, the discretization scheme accesses the corresponding columns of the new weight matrix at each iteration. 

This formulation allows us to efficiently compute the weights at each time step without explicitly storing all time-dependent weights, reducing memory usage.
We can reframe the neural network as~\eqref{eq:NewNetwork} in the discretize-then-optimize approach because we know the time points at which the network will be evaluated a priori, rather than in the neural ODE setting, where the time points are not fixed.
In this work, we discretize both the layers and weights using a forward Euler time integrator, followed by optimization using either ADAM or GNvpro.

\subsection{Optimize-then-discretize --- weight parameterization}\label{subsect:optdisc_weightParam}
For a neural ODE with constant weights, as defined by~\cite{ChenEtAl2018}, weight parameterization is necessary to compute the gradient of the objective function.  
This is because in the standard neural ODE formulation, weights are typically fixed over time, but for the system to evolve smoothly over time and for its gradients to be computed, the weights must be parameterized to allow the neural network to be trained using standard optimization techniques.

For time-dependent weights, we can still use the same parameterization as in the discretize-then-optimize approach, but we need to adapt the training process to account for the time-varying nature of the weights.
Once the weights are parameterized, they are truncated into a finite dimensional space, but rather than directly differentiating with respect to the original weights $\bftheta$, the network is trained to optimize the parameters $\bftheta_{\rm NODE}(t) = \{\bftheta_1, \dots, \bftheta_N\}$,  which represent the coefficients for the basis functions evaluated at different time points.

Recall that the adjoint ODE from section~\ref{sect:optdisc} governs the evolution of the adjoint variables $\mathbf{a}(t)$, which are used to compute the gradients of the loss function with respect to the weights. 
With the parameterized weights, we can rewrite the gradient of the loss function with respect to the time-dependent weights $\bftheta_P(t)$ using the chain rule: 
\begin{equation*}
    \frac{\partial \ell}{\partial \bftheta_P} = \frac{\partial \ell}{\partial \bfu(t)} \cdot \frac{\partial \bfu(t)}{\partial \bftheta_P(t)}.
\end{equation*}
Here, $\frac{\partial \ell}{\partial \bfu(t)}$ is the adjoint variable $\mathbf{a}(t)$, which represents the gradient of the loss function with respect to the state $\bfu(t)$, and $\frac{\partial \bfu(t)}{\partial \bftheta_P(t)}$ is the derivative of the state with respect to the parameterized weights $\bftheta_P(t)$.

Since $ \bftheta_P(t) $ is a function of polynomials $ p_i(t) $, we can rewrite  $\frac{\partial \bfu(t)}{\partial \bftheta_P(t)}$ as a linear combination of the derivatives of the state with respect to the basis functions $p_i(t)$:
\begin{equation}
    \frac{\partial \bfu(t)}{\partial \bftheta_P(t)} = \sum_{i=1}^{d} p_i(t) \frac{\partial \bfu(t)}{\partial \bftheta_i}\label{eq:gradient_param}.
\end{equation}

Now, instead of differentiating directly with respect to the original weights $\bftheta_i$, we need to project the gradient of the objective function onto the basis functions $p_i(t)$. 
This projection ensures that we are updating the parameters $ \bftheta_i $ in the direction that corresponds to the gradient of the loss function with respect to the parameterized weights $\bftheta_P(t)$.
This can be written as the integral 
\begin{equation}
    \frac{d \ell}{d \bftheta_i} = \int_{t_0}^{T} {\mathbf{a}(t)}^T \cdot p_i(t) \cdot \frac{\partial f(\bfu(t), t, \bftheta_P(t))}{\partial \bftheta_P(t)}  \, dt
\end{equation}
The term $\frac{\partial f(\bfu(t), t, \bftheta_P(t))}{\partial \bftheta_P(t)} $ computes how the hidden states of the network evolve with respect to the parameterized weights, and $\mathbf{a}(t)$ represents the adjoint variables.

Once we have the projected grad ient, the backpropagation process updates the coefficients $\bftheta_i$ by applying the gradient descent rule:
\begin{equation}
    \bftheta_i^{\text{new}} = \bftheta_i^{\text{old}} - \eta \frac{d \ell}{d \bftheta_i}
\end{equation}
where $\eta$ is the learning rate.
Thus, the gradients are applied to the parameterized weights, not directly to the original weights.

\subsection{Weight parameterization and stability}\label{subsect:weight_param_stability}
Weight parameterization plays a crucial role in ensuring the stability of the ODE system.
Parameterizing the weights in time provides a mechanism to control how rapidly the weights change, which ensures that the Jacobian matrices of the neural network layers, given by 
\begin{equation*}
    \mathbf{J}_1(t) = \frac{\partial f(\bfu(t), t, \bftheta_P(t))}{\partial \bftheta_P(t)} \;\; \text{and} \;\; \mathbf{J}_2(t) = \frac{\partial f(\bfu(t), t, \bftheta_P(t))}{\partial \bfu(t)}.
\end{equation*}
 evolve smoothly over time.

The spectrum of the Jacobian directly influences the behavior of the gradient.
In particular, limiting the rate of change in 
$\mathbf{J}(t)$ guarantees that its eigenvalues and eigenvectors change sufficiently slowly.
This smoothness helps prevent instability in the dynamics of the system~\cite{HaberRuthotto2017}.

It is also important to consider the stability of the polynomial basis functions used for parameterization.
The choice of polynomial basis functions can significantly affect the conditioning of the Vandermonde matrix, which is a matrix whose rows consist of the terms of a geometric progression, often in terms of powers of a variable $t_i$.
The condition number of the Vandermonde matrix, which quantifies the sensitivity of the matrix inversion process to numerical errors, plays a key role in determining the stability of the basis functions.

For a monomial basis function, the basis functions are of the form $\{1,t,t^2,\ldots,t^n \}$ and the corresponding Vandermonde matrix, where the rows correspond to the evaluation of the monomials at different time points is given by:
\begin{equation}
    \mathbf{V}_M =  \begin{pmatrix}
1 & t_0 & \cdots & t_0^n\\
1 & t_1 & \cdots & t_1^n\\
1 & t_2 & \cdots & t_2^n\\
\vdots & & \ddots & \\
1 & t_N & \cdots & t_N^n\\
\end{pmatrix}.
\end{equation}
Here, the system to find the weights $\bftheta_P(t)$ is 
\begin{equation}
   \mathbf{V}_M \cdot
\begin{pmatrix}
\bftheta_1^{(1)} \\ \bftheta_1^{(2)} \\ \bftheta_1^{(3)}\\\vdots\\\bftheta_1^{(n)}
\end{pmatrix}
=
\begin{pmatrix}
{\bftheta_P(t_0)}^{(1)} \\ {\bftheta_P(t_1)}^{(1)} \\ {\bftheta_P(t_2)}^{(1)} \\\vdots\\ {\bftheta_P(t_N)}^{(1)}
\end{pmatrix}.
\label{eq:Vandermonde_mono}
\end{equation}
This system can be ill-conditioned, meaning that small errors in the evaluation of the monomial at each time point can cause large errors in the approximation of the weights.
This is because the rows of the Vandermonde matrix for monomials are highly correlated for large $n$ and the condition number of $\mathbf{V}_M$, which measures the sensitivity of the solution to numerical perturbations, grows exponentially with the degree of the monomial.
This instability increases as $n$ increases, especially for a large number of evaluations $N$.

On the other hand, the Legendre polynomials $P_n(t)$ are defined over the interval $[-1,1]$ and have more favorable numerical properties. 
The Legendre polynomials are orthogonal, meaning that the inner product of any two distinct Legendre polynomials is zero.
This orthogonality helps reduce the correlation between the rows of the Vandermonde matrix, making it better conditioned than the monomial basis. 
The corresponding Vandermonde matrix for the Legendre basis is:
\begin{equation}
    \mathbf{V}_L =  \begin{pmatrix}
P_0(t_0) & P_1(t_0) & \cdots & P_n(t_0)\\
P_0(t_1) & P_1(t_1) & \cdots & P_n(t_1)\\
P_0(t_2) & P_1(t_2) & \cdots & P_n(t_2)\\
\vdots & & \ddots & \\
P_0(t_N) & P_1(t_N) & \cdots & P_n(t_N)\\
\end{pmatrix}
\end{equation}
where $P_i(t)$ are the Legendre polynomials evaluated at the time points.

Due to the orthogonality of the Legendre polynomials, the rows of the Vandermonde matrix $\mathbf{V}_L $ are much less correlated, resulting in a better-conditioned system.
The condition number does not grow as rapidly with $n$, and thus the system remains numerically stable even for larger degrees.

\begin{figure}[H]
    \centering
    \includegraphics[scale=0.8]{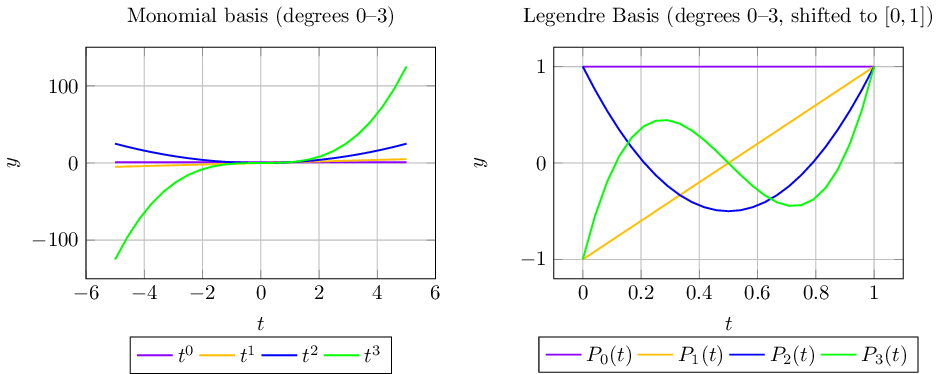}
    \caption{Comparison of monomial and shifted Legendre basis functions over $t \in [0,1]$ where each basis spans the space of degree-3 polynomials but with different conditioning and orthogonality properties}\label{fig:polynomials}
\end{figure}

The stability of the Legendre basis is particularly beneficial when training neural networks with parameterized weights because it ensures that the weights do not change erratically during training, allowing for faster convergence and better generalization. 
Conversely, the instability of the monomial basis can cause slow or even no convergence due to the numerical challenges associated with solving the ill-conditioned system.
This is demonstrated in section~\ref{sect:results}.
A visualization of each polyomial basis function for degrees $0$ through $3$ is presented in figure~\ref{fig:polynomials}.

\section{Computational results}\label{sect:results}
In this section, we outline the methods used in our experiments and interpret the results of implementing weight parameterization for neural ODEs and ResNets on three surrogate modeling tasks.
Traditional ResNet architectures generally do not employ weight parameterization, but omitting this feature can introduce unnecessary computational burden due to an excess of function evaluations or high-dimensional weight matrices.
On the other hand, while neural ODEs traditionally require parameterization to compute gradients, they do not always use time-dependent weights, $t$ as an input feature, or polynomial basis function to parameterize weights in time.
Thus, our goal is to investigate the effectiveness of polynomial weight parameterization for improving performance and efficiency in surrogate modeling tasks.

We focus on the comparison between parameterized and non-parameterized networks, examining both discretize-then-optimize and optimize-then-discretize training methods.
We also evaluate how the choice of polynomial basis influences model performance.
We aim to clarify under which circumstances weight parameterization is the most useful, and how different choices regarding the polynomial basis function can impact the outcome. 
Our main findings are:
\begin{itemize}
    \item A third-order monomial parameterization consistently fails to converge to the desired loss tolerance and is not conducive to efficiency and error reduction in ResNets.
    \item Equipping a ResNet with a Legendre polynomial basis achieves similar accuracy and loss as its non-parameterized counterpart while maintaining a reduction of weights. 
    \item Legendre parameterization requires much fewer function evaluations than monomial parameterization due to the adaptive time integrator in a neural ODE.\@
    \item A third order Legendre-parameterized neural ODE attains a lower training loss than an analogous non-parameterized neural ODE.\@
    \item Legendre parameterization increases the amount of network weights (and thus expressivity) in a neural ODE with less computing time and cost.
\end{itemize}

We compare two polynomial bases for parameterization: a monomial basis and a Legendre polynomial basis. 
The Legendre basis is orthogonal by construction, in contrast to the monomial basis, which allows us to isolate the effect of orthogonality.
We primarily focus on polynomials of degree 3 but also investigate higher-order terms to evaluate the trade-off between expressiveness and computational cost.

\subsection{Datasets}\label{subsect:data}

Our study involves three surrogate modeling problems:
\begin{itemize}
    \item Energy Exascale Earth System Model Land Model (ELM): a 15-parameter model with 10 output quantities. The dataset contains 1740 training, 249 test, and 497 validation points.
    \item CDR (Convection-Diffusion-Reaction): A PDE system with $55\times800$ parameter samples and $72\times800$ output targets.
    \item DCR (Diffusion-Convection): A Poisson-type PDE with $3\times10,000$ inputs and $882\times10,000$ outputs.
\end{itemize}

All experiments use the MATLAB Meganet library~\cite{HaberRuthotto2017} and PyTorch.
For each dataset, we compare the performance of parameterized neural networks as surrogates to their non-parameterize analogues, and the potential benefits of weight parameterization in terms of training time, convergence, and accuracy.
The CDR dataset is given by the system of equations
\begin{align*}
     du/dt &= \nabla \cdot (D \nabla u) - v \cdot \nabla u + f + y' * r(u) \\ 
     D \nabla u \cdot n &= 0 \quad \text{(Neumann boundary conditions)} \\
     u &= 0 \quad \text{(initial condition)}
\end{align*}
for state variable $u$, parameters $y$, and others.
Here, $y$ are parameters $55 \times 800$ and $c$ are targets $72 \times 800$.
The DCR model is given by 
\begin{align*}
     -\nabla \cdot (m(x;y) \nabla u) &= q  \quad   \text{(Poisson's Equation)}\\
    \nabla u \cdot n &= 0 \quad \text{(Neumann boundary conditions)}                  
\end{align*}
where $y$ are parameters $3 \times 10000$ and $c$ are targets $882 \times 10000$.
More detailed information on the DCR and CDR models can be found in~\cite{VarPro}.

\subsection{ResNet results}
We evaluate the impact of weight parameterization
on two principle architectures: a ResNet with a forward Euler time integrator, and a Hamiltonian-inspired network with a Verlet time integrator~\cite{HaberRuthotto2017}, particularly due to their ubiquity in the pre-existing literature and superior performance against a ResNet with a Runge-Kutta 4 time integrator and a Leapfrog architecture (a special case of the Hamiltonian network).

Across all three surrogate modeling tasks—ELM, CDR, and DCR—we compare third-order monomial and Legendre polynomial basis parameterizations against non-parameterized baselines. 
Training is performed using both ADAM and GNvpro optimizers over a maximum of 1000 epochs, with a learning rate of 0.001 and batch size of 32.
We ran each surrogate example over $12$ time steps with $15$ channels.

\begin{figure}[H]
    \centering
    \includegraphics[scale=0.65]{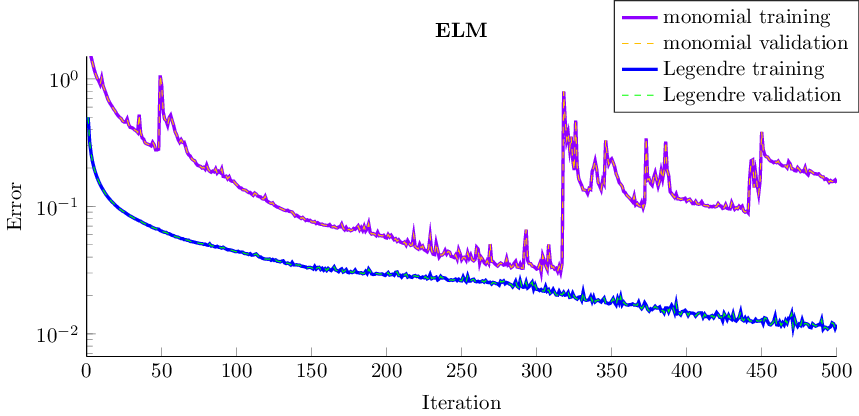}
    \includegraphics[scale=0.65]{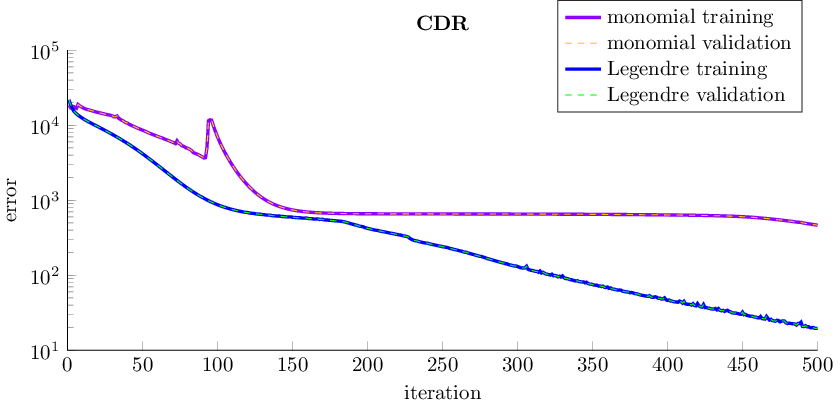}
    \includegraphics[scale=0.65]{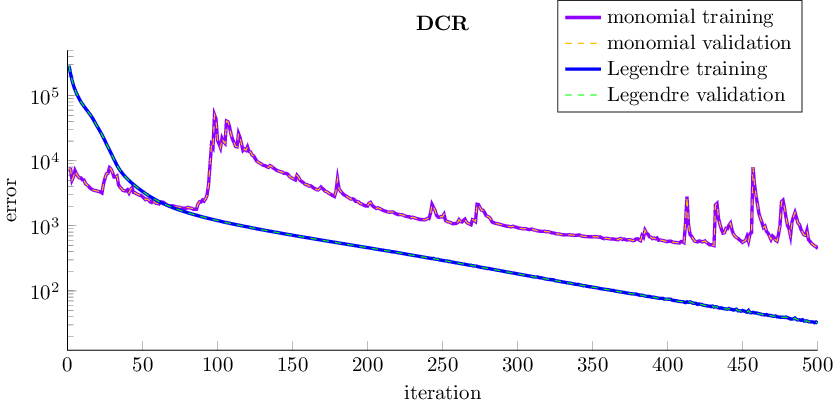}
    \caption{Training and testing error attained by a third degree Legendre-and monomial-parameterized ResNet using the ADAM optimization algorithm}\label{fig:ResNet_Loss}
\end{figure}

Figure~\ref{fig:ResNet_Loss} presents the training and testing error curves for ResNets trained on each surrogate task. 
Across all cases, third-degree Legendre parameterization consistently outperforms monomial parameterization, demonstrating faster convergence and lower errors. 
Monomial-parameterized networks often stall or converge poorly, particularly in the DCR example, where both train and test errors remain high. 
This confirms the numerical instability and poor conditioning of the monomial Vandermonde matrix, as discussed in section~\ref{subsect:weight_param_stability}.

\begin{figure}[H]
    \centering
    \includegraphics[scale=0.65]{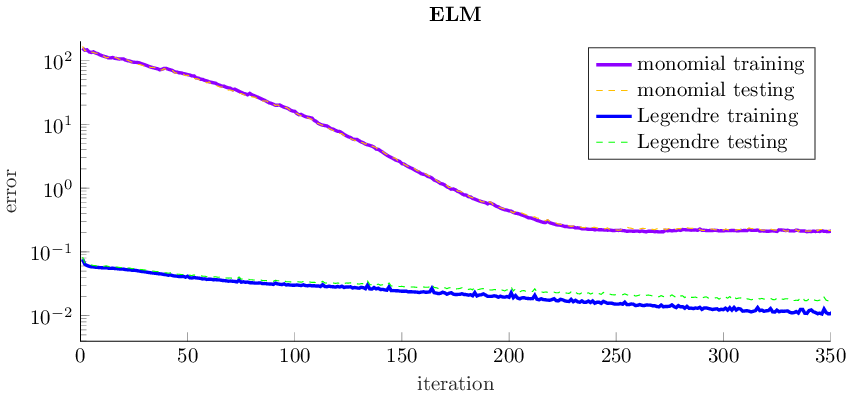}
    \includegraphics[scale=0.65]{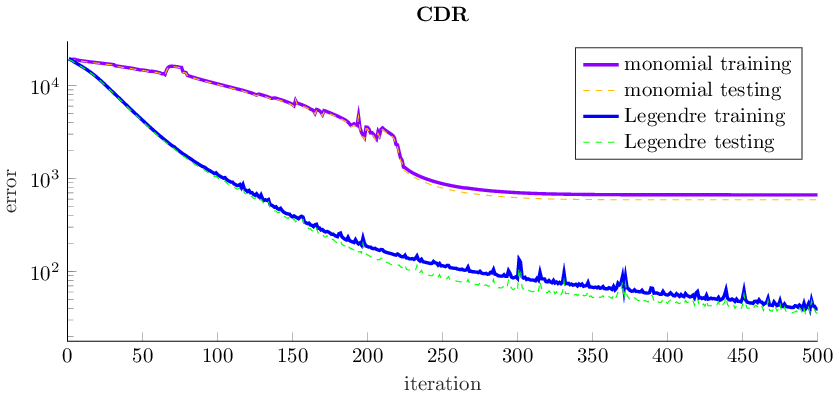}
    \includegraphics[scale=0.65]{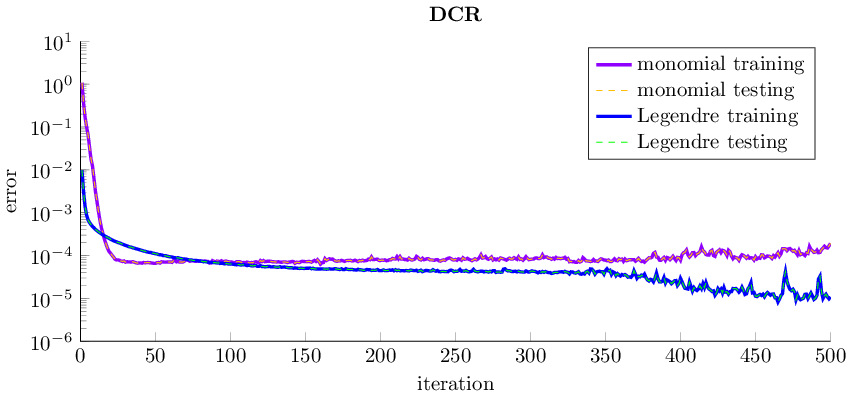}
    \caption{Training and testing error attained by a third degree Legendre-and monomial-parameterized Hamiltonian network using the ADAM optimization algorithm}\label{fig:Hamiltonian_Loss}
\end{figure}

Figure~\ref{fig:Hamiltonian_Loss} illustrates results for Hamiltonian-inspired ResNets. 
Here, Legendre parameterization also yields lower errors and faster convergence than monomial parameterization for all datasets. 
As such, this architecture appears particularly well-suited for polynomial-parameterized weights. 
Notably, in the ELM and CDR cases, the Legendre-Hamiltonian networks match or outperform their non-parameterized counterparts, suggesting that Legendre parameterization introduces beneficial regularization and structure, particularly when paired with the energy-conserving properties of Hamiltonian dynamics.

%

Table~\ref{tab:ComprehensiveLegendre} provides a detailed breakdown of training loss across architectures and depths (denoted by $T$).
For ELM and DCR, we observe that Legendre parameterized networks generally achieve comparable or improved loss values relative to non-parameterized versions, especially in shallower networks. 
For example, in the Hamiltonian case with 
$T=1$, Legendre parameterization yields a lower loss (0.0075) than the non-parameterized baseline (0.0079), as shown in table~\ref{tab:ParamvsNonDiscOpt}. 
However, for deeper networks (e.g., $T=10$), this benefit diminishes, and in some cases, performance worsens slightly (e.g., Legendre loss = 0.0257 vs.\ baseline = 0.0159). 
This suggests a depth-dependent trade-off where the expressivity gains of parameterization must be balanced against overfitting and the increasing complexity of learning in higher-dimensional parameter spaces.

\begin{figure}[H]
    \centering
    \includegraphics[scale=0.75]{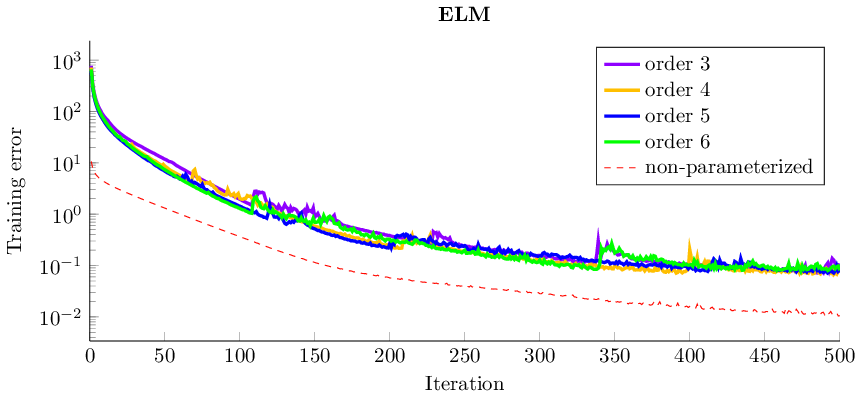}
    \includegraphics[scale=0.75]{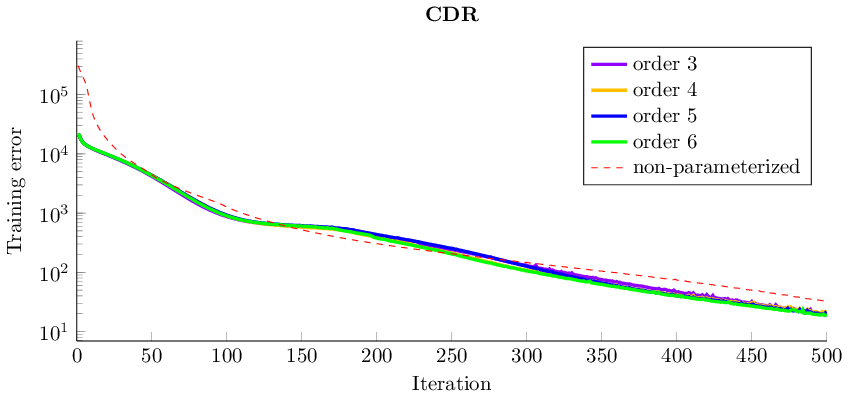}
    \includegraphics[scale=0.75]{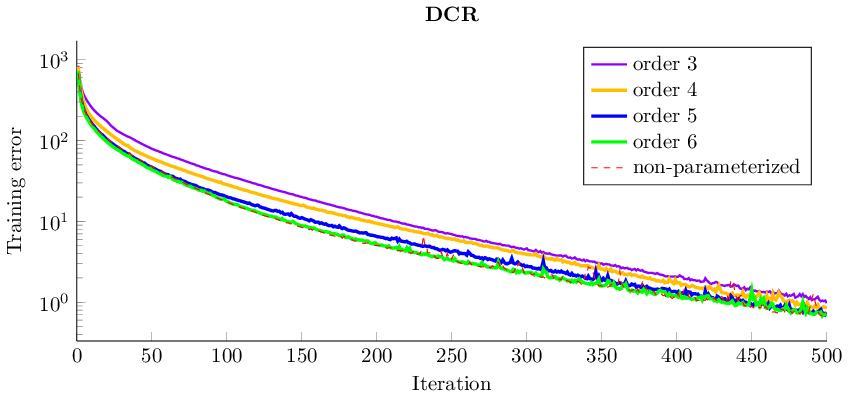}
    \caption{Convergence of training error across iterations for a Legendre-parameterized ResNet of increasing degree using the ADAM optimization algorithm}\label{fig:comparing degrees ResNet}
\end{figure}

\begin{figure}[H]
    \centering
        \includegraphics[scale=0.75]{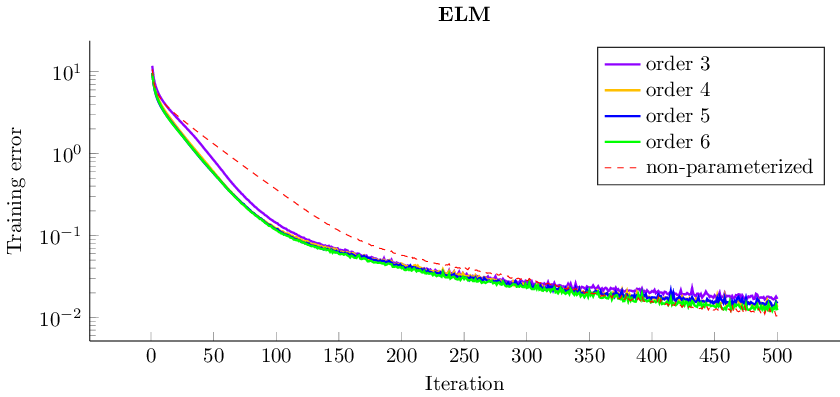}
        \includegraphics[scale=0.75]{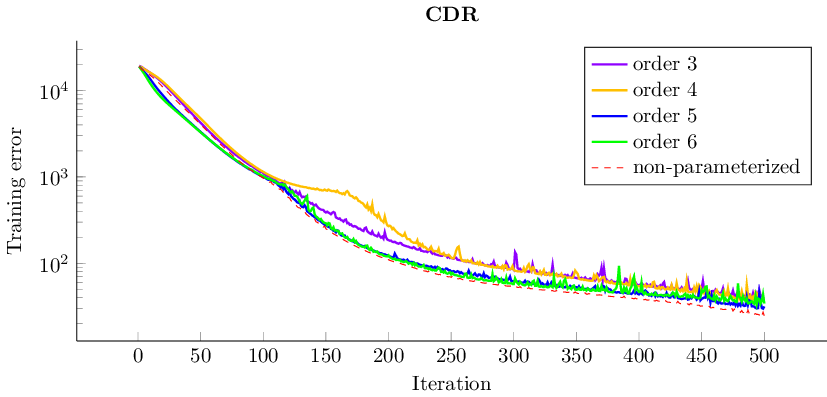}
        \includegraphics[scale=0.75]{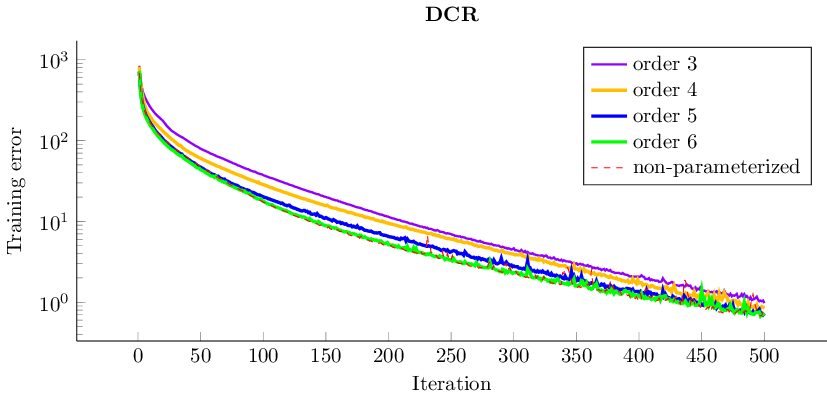}
        \caption{Convergence of training error across iterations for a Legendre-parameterized Hamiltonian network of increasing degree using the ADAM optimization algorithm}\label{fig:comparing degrees Ham}
\end{figure}

Table~\ref{tab:DegofFreedom} explores this trade-off in more detail for the DCR problem under a Hamiltonian architecture. 
As the degree of the Legendre polynomial basis increases from 3 to 6, the training error steadily decreases, even becoming lower than the non-parameterized baseline at degree 6. 
However, this comes at the cost of increased degrees of freedom, from 1005 to 1725 weights, which may lead to overfitting in low-data regimes. 
This trade-off is further depicted in figure~\ref{fig:comparing degrees ResNet}, which shows convergence curves for different Legendre polynomial orders: as the order increases, training and testing curves more closely resemble those of the non-parameterized networks.
Figure~\ref{fig:comparing degrees Ham} extends this comparison to Hamiltonian networks. 
The same trend holds: higher-order Legendre parameterizations produce training/test errors that approach the non-parameterized performance, with degree-6 curves nearly indistinguishable from the baseline in some cases. 
This demonstrates that the expressivity of the Legendre basis can compensate for the loss of flexibility due to parameter reduction, though only up to a point.

Taken together, these results support the use of Legendre parameterization as a computationally efficient alternative to non-parameterized weights in ResNets, particularly when paired with Hamiltonian architectures.
The reduced number of parameters improves storage and training efficiency while maintaining competitive performance. 
Moreover, the Hamiltonian structure appears to enhance the stability of parameterized networks, especially when using orthogonal bases like Legendre, likely due to the geometric preservation of energy dynamics via the Verlet integrator.

\subsection{Neural ODE results}
We now discuss the effect of weight parameterization for neural ODEs trained on the same three surrogate modeling problems: ELM, CDR, and DCR.\@ In this setting, weights are parameterized as time-dependent functions using either third-order monomial or Legendre polynomial bases. The continuous model is trained using the optimize-then-discretize framework and solved numerically with the Dormand-Prince (DOPRI5) method from SciPy, an adaptive 4(5) Runge-Kutta integrator.

Table~\ref{tab:meanstd} presents the mean and standard deviation of training and validation errors over five random seeds, as well as the total number of function evaluations required for convergence. 
Across all datasets, Legendre parameterization yields significantly fewer function evaluations, often by an order of magnitude, compared to monomial parameterization, despite achieving similar or better final error. 
This illustrates the clear advantage of orthogonal polynomial bases when used with adaptive time integration. 
Specifically, for the DCR example, the Legendre-parameterized neural ODE converges in roughly $9.17\times10^7$ function evaluations, compared to over $7.87\times10^8$ function evaluations for the monomial case, offering nearly an $8\times$ improvement in computational efficiency.

\begin{figure}[H]
    \centering
        \includegraphics[scale=0.75]{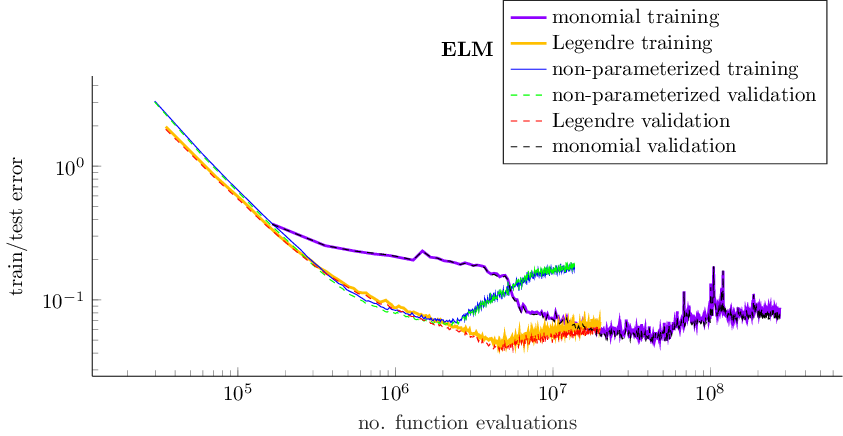}
        \includegraphics[scale=0.75]{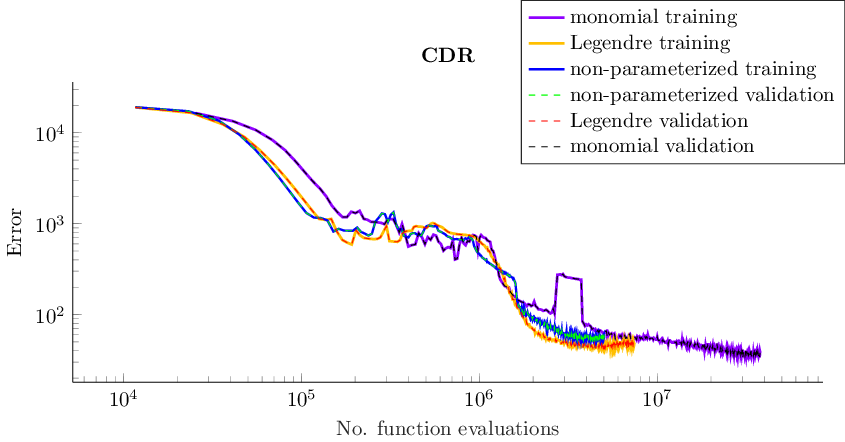}       
         \includegraphics[scale=0.75]{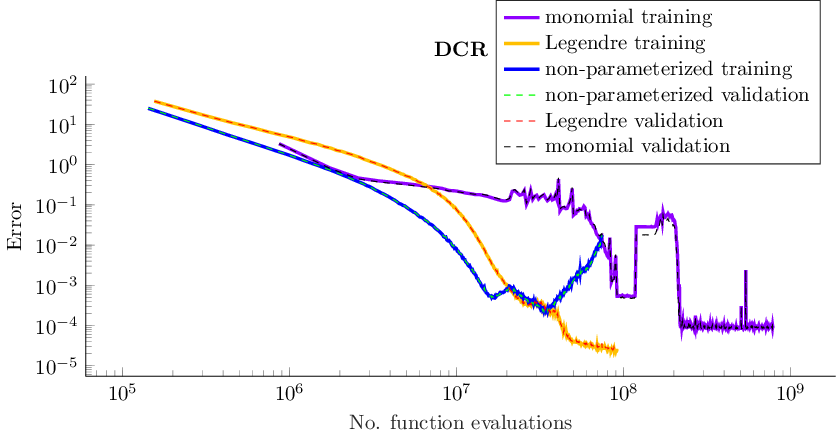}
        \caption{Neural ODE training and validation loss compared to number of function evaluations for the surrogate examples for a time-span of $T = [0,1]$}\label{fig:neuralODE_loss}
\end{figure}

Figure~\ref{fig:neuralODE_loss} visualizes the training and validation error for the neural ODE versus the number of function evaluations. 
In all three surrogate modeling tasks, the Legendre parameterized network converges faster and to lower or comparable loss than both the monomial and non-parameterized networks. 
For the DCR surrogate, the Legendre model reaches a validation loss of approximately $0.28$ after $10^8$ evaluations, whereas the monomial case fails to drop below $0.6$ despite nearly $8\times$ more compute.
In the ELM example, Legendre achieves lower loss than the non-parameterized model and outperforms monomial consistently.
For CDR, all models eventually converge, but Legendre remains the most efficient, exhibiting smoother error reduction.

These results confirm several key insights; Legendre parameterization introduces negligible overhead in terms of implementation or FLOPs, yet substantially reduces cost via fewer function evaluations.
The orthogonality of Legendre polynomials improves numerical conditioning of the parameterization and reduces instability in the adjoint method's backward pass.
In contrast, monomial parameterization suffers from ill-conditioning, leading to unstable or inefficient training, despite having the same degree and expressive capacity.

Higher-degree Legendre and monomial basis results from the ResNet experiments suggest that extending these tests to degrees 4–6 would likely yield similar trends: increased expressiveness at the cost of more parameters, but still more efficient than monomials.
Together, these findings illustrate that Legendre-based weight parameterization enables faster, more stable training of neural ODEs without sacrificing accuracy, making it a strong candidate for surrogate modeling tasks where computational efficiency and model interpretability are paramount.

\section{Discussion}\label{sect:discussion}
In this work, we have studied how weight parameterization impacts ResNets and neural ODEs for surrogate modeling tasks, both of which require unique considerations for weight parameterization based on their respective training algorithms.
Hence, we elucidate potential benefits of weight parameterization compared to non-parameterized approaches in the context of both discretize-then-optimize and optimize-then-discretize training methods.
We also investigate which neural network dynamics and parameters are most conducive to the helpfulness of weight parameterization, and the influence of different choices and orders of basis functions on the outcome.

In the discretize-then-optimize sense, we find that weight parameterization has the potential to reduce weights while maintaining effective performance of a ResNet depending on the choice of basis function.
For each surrogate modeling task, we note that a third-order monomial parameterization consistently fails to converge to the desired loss tolerance and conclude that this choice of basis function does not improve efficiency nor loss minimization of the ResNet.

On the other hand, a ResNet parameterized with a Legendre polynomial basis achieves similar accuracy and loss as its non-parameterized analogue while maintaining a reduction of weights. 
This weight reduction is particularly helpful in mitigating the computational expense of ResNets since the amount of weights typically varies at each layer.
Although the cost of implementing a Legendre and monomial parameterization is the same, the Legendre basis reduces the number of weights without much loss of accuracy and is ultimately the preferred choice for weight parameterization.

For a neural ODE, the computational cost of interpolating in time is independent of the number of training samples, since the interpolation cost is computed once and applied to the entire batch. 
Hence, adding weights doesn't significantly alter the computational expense in terms of FLOPs, and we are more concerned with the number of function evaluations.
Adding weights, however, does provide higher expressiveness of the neural ODE~\cite{autonomous} and thus a lower error overall.

We see from the results that while the order $3$ monomial and Legendre basis functions are equally expressive, they require a vastly different amount of function evaluations to converge. 
It is clear from the results that a Legendre parameterization is cheaper to implement due to the adaptive time integrator. 
Furthermore, parameterizing with respect to a Legendre basis function only marginally increases the cost of evaluating the neural ODE while drastically reducing the number of function evaluations and improving training and validation error.

We see in both ResNets and neural ODEs that a monomial basis of degree 3 can impede convergence, suggesting that higher expressiveness does not always predicate a lower error.
This suggests that orthogonality of the polynomial basis function plays a key role in the success of weight parameterization methods.
By studying multiple combinations of training algorithms, architectures, and polynomial bases, we observe that the orthogonal Legendre basis function, consistently outperforms the monomial basis.
The Legendre basis demonstrates stability in its performance across changes in network depth, architecture, and optimization algorithm. 
For ResNets, the Legendre basis particularly works well with the Hamiltonian-inspired architecture, likely due to the energy-conserving nature of the Verlet integrator and its compatibility with smooth weight evolution.
Our results indicate that Hamiltonian-inspired architectures synergize well with orthogonal polynomial parameterizations. 
In conclusion, for both the ResNet and neural ODE, weight parameterization offers potential benefits in computational cost and accuracy, but in our three test cases and ODE task, the success of this method markedly depends on the choice of basis function.

\section{Future work}\label{sect:future}
The results presented in this work demonstrate that polynomial weight parameterization, particularly with orthogonal bases such as Legendre polynomials, offers significant advantages in terms of training efficiency, expressivity, and stability across a variety of continuous-time neural network architectures. 
These findings suggest several avenues for future work, both theoretical and applied.

While this work focuses on polynomial bases, alternative parameterizations such as Fourier series, splines, or neural network-based time embeddings may offer richer representations for capturing periodic, localized, or highly nonlinear temporal dynamics. 
This time-dependent weight parameterization framework offers a continuous-time network design in which architectural complexity adapts smoothly over time.
Using this architecture, one could, for instance, investigate sparse or low-rank weight structures that vary in time, enabling efficient real-time control, adaptation, or resource-aware inference.
This could be especially beneficial in many real-world applications.

Polynomial-parameterized continuous-time networks are well-suited for surrogate modeling in scientific computing, especially where data is scarce and interpretability is essential. 
In particular, domains such as climate modeling, epidemiology, physics-informed machine learning, and inverse problems in imaging could benefit from compact, expressive, and stable models. 
Additionally, the low-parameter regime for ResNets enabled by this approach makes it attractive for scenarios with strict memory or computation constraints.

In summary, time-based weight parameterization provides versatility and efficiency for building continuous-time deep neural network models. 
While we demonstrate the benefits of this approach in the context of surrogate modeling tasks, its potential extends to a wide range of applications in machine learning and scientific computing, offering promising directions for both methodological innovation and practical deployment in complex real-world systems.

\newpage

\section{Tables}\label{sect:tables}

\begin{table}[h]
    \centering
    \caption{An overview of components found in different pieces of broader neural ODE literature, and how our formulation and studies compare.}\label{tab:literature}
    \begin{tabularx}{\textwidth}{lcccccc}
        \toprule
        paper & \makecell{time-dependent\\weights} & \makecell{\hspace{0.1cm}parameterized\\weights} & \makecell{optimize-\\discretize} & \makecell{discretize-\\optimize} & \makecell{orthogonal\\weights} \\
        \midrule
        Chen et.~al.~\cite{ChenEtAl2018} & \xmark~& \xmark~& \cmark~& \xmark~& \xmark~\\ 
        Davis et.~al.~\cite{autonomous} & \cmark~& \cmark~& \cmark~& \xmark~& \cmark~\\
        Gunther et.~al.~\cite{gunther} & \cmark~& \cmark~& \xmark~& \cmark~& \xmark~\\ 
        Massaroli et.~al.~\cite{dissect} & \cmark~& \cmark~& \cmark~& \xmark~& \cmark~\\
        Yu et.~al.~\cite{generalized} & \cmark~& \cmark~& \xmark~& \cmark~& \xmark~\\ 
        Ott et.~al.~\cite{ott2021resnet} & \xmark~& \xmark~& \cmark~& \xmark~& \xmark~\\
        Zhou et.~al.~\cite{zhouNODES} & \cmark~& \cmark~& \xmark~& \cmark~& \xmark~\\ 
        Yu et.~al.~\cite{Huang_Liu_Lang_Yu_Wang_Li_2018} & \cmark~& \cmark~& \xmark~& \cmark~& \xmark~\\
        \textbf{ours} & \cmark~& \cmark~& \cmark~& \cmark~& \cmark~\\
        \botrule%
    \end{tabularx}%
\end{table}

\begin{table}[h]
    \caption{This table compares the training loss attained at different depths by the ELM surrogate model across different network architectures and values of $T$. These networks are trained using GNvpro and parameterized with a third-order Legendre basis. The bottom row of each set of $T$ for each data set corresponds to the non-parameterized value.}\label{tab:ComprehensiveLegendre}
    \centering
    \begin{tabular}{llccccccc}
        \toprule
        & & & \multicolumn{2}{c}{ResNet} & \multicolumn{2}{c}{Hamiltonian} & \multicolumn{2}{c}{Leapfrog} \\
        \cmidrule(lr){4-5} \cmidrule(lr){6-7} \cmidrule(lr){8-9}
        Dataset & Parameterization & Depth & ADAM & GNvpro & ADAM & GNvpro & ADAM & GNvpro \\
        \midrule
        \multirow{6}{*}{ELM} & \multirow{3}{*}{Legendre ($d=3$)} & $T=1$ & 0.0199 & 0.0080 & 0.0193 & 0.0075 & 0.0242 & 0.0086 \\
        & & $T=5$ & 0.0332 & 0.0079 & \textbf{0.0173} & \textbf{0.0037} & 0.0184 & 0.0055 \\ 
        & & $T=10$ & 0.0719 & \textbf{0.0208} & \textbf{0.0166} & 0.0257 & 0.0173 & 0.0396\\ 
        \cmidrule(lr){2-9}
        & \multirow{3}{*}{none} & $T=1$ & 0.0202 & 0.0088 & 0.0215 & 0.0079 & 0.0221 & 0.0082 \\
        & & $T=5$ & 0.0203 & 0.0045 & 0.0123 & 0.0023 & 0.0144 & 0.0028 \\
        & & $T=10$ & 0.0616 & 0.0164 & 0.0104 & 0.0159 & 0.0361 & 0.0283 \\
        \midrule
        \multirow{6}{*}{DCR} & \multirow{3}{*}{Legendre ($d=3$)} & $T=1$ & 2.9206 & 6.6563e-06 & 0.4047 & 5.1356e-06 & 0.3044 & 6.8228e-06 \\
        & & $T=5$ & 9.0959 & 2.8587e-06 & 0.1868 & 5.1356e-06 & 0.1781 & 5.6372e-06 \\
        & & $T=10$ & 32.0303 & 2.5656e-06 & 0.9789 & 6.5758e-06 & 9.3198 & 6.6189e-06 \\
        \cmidrule(lr){2-9}
        & \multirow{3}{*}{none} & $T=1$ & 3.7591 & 6.7030e-06 & 0.5189 & 5.1356e-06 & 0.3875 & 6.9385e-06 \\
        & & $T=5$ & 8.7636 & 6.5177e-06 & 0.1284 & 6.6983e-06 & 0.1420 & 6.7277e-06 \\
        & & $T=10$ & 32.1492 & 6.6763e-06 & 0.6806 & 6.7587e-06 & 5.1977 & 6.6647e-06 \\
        \midrule
        \multirow{6}{*}{CDR} & \multirow{3}{*}{Legendre ($d=3$)} & $T=1$ & 0.0242 & 0.0184 & 0.0173 & 0.2528 & 71.9807 & 0.0444 \\
        & & $T=5$ & 0.0086 & 0.0055 & 0.0396 & 1.1313 & 19.1215 & 0.0493 \\
        & & $T=10$ & 19.5592 & 0.9343 & 39.4597 & 150.8030 & 34.8918 & 2.3839 \\
        \cmidrule(lr){2-9}
        & \multirow{3}{*}{none} & $T=1$ & 634 & 0.1969 & 633.9463 & 0.2288 & 89.1428 & 0.0575 \\
        & & $T=5$ & 25.1991 & 0.0697 & 24.5704 & 0.0670 & 23.0403 & 0.0497 \\
        & & $T=10$ & 15.8598 & 1.0428 & 24.5564 & 7.8833 & 34.7752 & 0.1622 \\
        \bottomrule
    \end{tabular}
\end{table}

\begin{table}[h]
 \caption{This table compares the training loss attained at different depths by the ELM surrogate model for a parameterized and non-parameterized case. The network is Hamiltonian-inspired and parameterized with a third-order Legendre polynomial basis.}\label{tab:ParamvsNonDiscOpt}
\begin{tabular}{@{}lllllll@{}}
\toprule
& \multicolumn{3}{c}{Gnvpro} & \multicolumn{3}{c}{ADAM} \\
\cmidrule{2-7}
  Depth & $T=1$ & $T=5$ & $T=10$ & $T=1$ & $T=5$ & $T=10$\\
\textcolor{red}{Non-Parameterized} & 0.0079 & 0.0023 & 0.0159 & 0.0215 & 0.0123 & 0.0104\\ 
Parameterized & 0.0075 & 0.0037 & 0.0257 & 0.0193 & 0.0173 & 0.0166\\  
\botrule%
\end{tabular}
\end{table}

\begin{table}[h]
    \caption{This table depicts the change in training error of the DCR surrogate model between the Legendre-Hamiltonian case and a non-parameterized network under the Hamiltonian architecture, optimized  using ADAM.}\label{tab:DegofFreedom}
    \centering
    \begin{tabular}{ccc}
    \toprule
         Order & $\Delta$ \textbf{Error}& Deg.\ of Freedom \\ \midrule
          $3$ & 0.2983 & 1005 \\
     $4$ & 0.1363 & 1245 \\
     $5$ & 0.0176 & 1485\\     
     $6$ & -0.0033  & 1725\\ 
    \botrule%
    \end{tabular}
\end{table}

\begin{table}
    \caption{This table displays the mean and standard deviation of the training and validation errors of the neural ODE computed by the optimize-then-discretize approach for all three surrogate examples. The two farthest-right columns are the number of training evaluations required to reach convergence for a third-order Legendre parameterization (left) and a third-order monomial parameterization (right). 
    }\label{tab:meanstd}
    \centering
    \begin{tabular}{llccccccc}
    \toprule
     \multicolumn{2}{c}{} & \multicolumn{2}{c}{\textbf{mean}}  & \multicolumn{2}{c}{\textbf{standard dev.}} & \multicolumn{2}{c}{\textbf{no.\ function evals}} \\
    & & Legendre & monomial & Legendre & monomial & Legendre & monomial \\
    \cmidrule(lr){3-4} \cmidrule(lr){5-6} \cmidrule(lr){7-8}
    \multirow{2}{*}{\textbf{ELM}} & training & 0.066 & 0.078 & 0.022 & 0.01 & \multirow{2}{*}{$1.99 \times 10^7$} & \multirow{2}{*}{$2.75 \times 10^8$}\\
    & validation & 0.071 & 0.081 & 0.099 & 0.042 & &\\
    \midrule
    \multirow{2}{*}{\textbf{CDR}} & training & 282.2 & 264.9 & 104.015 & 26.452 & \multirow{2}{*}{$7.43 \times 10^6$} & \multirow{2}{*}{$3.78 \times 10^7$}\\
     & validation &326.4 &297.9 & 92.329 & 27.598 & &\\
     \midrule
    \multirow{2}{*}{\textbf{DCR}} & training & 0.282 & 0.036 & 0.617 & 0.0552 & \multirow{2}{*}{$9.17 \times 10^7$} & \multirow{2}{*}{$7.87 \times 10^8$}\\
     & validation & 0.281 & 0.037 & 0.607 & 0.056 & & \\
     \botrule%
\end{tabular}

\end{table}

\newpage

\section{Figures}\label{sect:Figures}

\section*{Statements and declarations}

\begin{itemize}
    \item Funding: L.R.'s and H.R.'s work was partially supported by the US National Science Foundation under grant DMS 2038118.
    \item Conflict of interest: The authors declare that they have no known competing financial interests or personal relationships that could have appeared to influence the work reported in this paper.
    \item Author contributions: H.R. wrote the main manuscript text, prepared figures and tables, and performed experiments and analysis of results. L.R. provided supervision, formed methodology,  and reviewed and edited the main manuscript text. K.S. conceptualized the project, formed methodology, and provided funding and technical resources.
    \item Data availability: The datasets used in this study are available upon reasonable request from the corresponding author. 
    The CDR dataset can be downloaded \href{https://math.emory.edu/~lruthot/publication/newman-et-al-2020/CDR_data.mat}{here}.\relax
    \, The CDR dataset can be downloaded \href{https://math.emory.edu/~lruthot/publication/newman-et-al-2020/DCR_data.mat}{here}.\relax
    \, All ResNet experiments were run using the \href{https://github.com/XtractOpen/Meganet.m}{MATLAB Meganet library} and PyTorch.
    \item Ethics approval: not applicable.
    \item Consent to participate: not applicable.
    \item Consent for publication: not applicable. 
\end{itemize}

This work is partially supported by Sandia National Laboratories' Laboratory Directed Research and Development (LDRD) program. This article has been co-authored by employees of National Technology \& Engineering Solutions of Sandia, LLC under Contract No. DE-NA0003525 with the U.S. Department of Energy (DOE). The employees co-own right, title and interest in and to the article and are responsible for its contents. The United States Government retains and the publisher, by accepting the article for publication, acknowledges that the United States Government retains a non-exclusive, paid-up, irrevocable, world-wide license to publish or reproduce the published form of this article or allow others to do so, for United States Government purposes. The DOE will provide public access to these results of federally sponsored research in accordance with the \href{https://www.energy.gov/downloads/doe-public-access-plan}{DOE Public Access Plan}. 

\bibliography{main}


\begin{thebibliography}{34}
\ifx \bisbn   \undefined \def \bisbn  #1{ISBN #1}\fi
\ifx \binits  \undefined \def \binits#1{#1}\fi
\ifx \bauthor  \undefined \def \bauthor#1{#1}\fi
\ifx \batitle  \undefined \def \batitle#1{#1}\fi
\ifx \bjtitle  \undefined \def \bjtitle#1{#1}\fi
\ifx \bvolume  \undefined \def \bvolume#1{\textbf{#1}}\fi
\ifx \byear  \undefined \def \byear#1{#1}\fi
\ifx \bissue  \undefined \def \bissue#1{#1}\fi
\ifx \bfpage  \undefined \def \bfpage#1{#1}\fi
\ifx \blpage  \undefined \def \blpage #1{#1}\fi
\ifx \burl  \undefined \def \burl#1{\textsf{#1}}\fi
\ifx \doiurl  \undefined \def \doiurl#1{\url{https://doi.org/#1}}\fi
\ifx \betal  \undefined \def \betal{\textit{et al.}}\fi
\ifx \binstitute  \undefined \def \binstitute#1{#1}\fi
\ifx \binstitutionaled  \undefined \def \binstitutionaled#1{#1}\fi
\ifx \bctitle  \undefined \def \bctitle#1{#1}\fi
\ifx \beditor  \undefined \def \beditor#1{#1}\fi
\ifx \bpublisher  \undefined \def \bpublisher#1{#1}\fi
\ifx \bbtitle  \undefined \def \bbtitle#1{#1}\fi
\ifx \bedition  \undefined \def \bedition#1{#1}\fi
\ifx \bseriesno  \undefined \def \bseriesno#1{#1}\fi
\ifx \blocation  \undefined \def \blocation#1{#1}\fi
\ifx \bsertitle  \undefined \def \bsertitle#1{#1}\fi
\ifx \bsnm \undefined \def \bsnm#1{#1}\fi
\ifx \bsuffix \undefined \def \bsuffix#1{#1}\fi
\ifx \bparticle \undefined \def \bparticle#1{#1}\fi
\ifx \barticle \undefined \def \barticle#1{#1}\fi
\bibcommenthead
\ifx \bconfdate \undefined \def \bconfdate #1{#1}\fi
\ifx \botherref \undefined \def \botherref #1{#1}\fi
\ifx \url \undefined \def \url#1{\textsf{#1}}\fi
\ifx \bchapter \undefined \def \bchapter#1{#1}\fi
\ifx \bbook \undefined \def \bbook#1{#1}\fi
\ifx \bcomment \undefined \def \bcomment#1{#1}\fi
\ifx \oauthor \undefined \def \oauthor#1{#1}\fi
\ifx \citeauthoryear \undefined \def \citeauthoryear#1{#1}\fi
\ifx \endbibitem  \undefined \def \endbibitem {}\fi
\ifx \bconflocation  \undefined \def \bconflocation#1{#1}\fi
\ifx \arxivurl  \undefined \def \arxivurl#1{\textsf{#1}}\fi
\csname PreBibitemsHook\endcsname

\bibitem[\protect\citeauthoryear{Sudret}{2019}]{Sudret2019}
\begin{bchapter}
\bauthor{\bsnm{Sudret}, \binits{B.}}:
\bctitle{Surrogate models for uncertainty quantification and design optimization}.
In: \bbtitle{Summer School of the German Research School for Simulation Sciences}
(\byear{2019}).
\doiurl{10.3929/ethz-b-000359599}
\end{bchapter}
\endbibitem

\bibitem[\protect\citeauthoryear{Kovachki et~al.}{2021}]{KovachkiTgCNN2021}
\begin{barticle}
\bauthor{\bsnm{Kovachki}, \binits{N.B.}}, \betal:
\batitle{Tgcnn: An efficient surrogate for real-time data assimilation in subsurface flow}.
\bjtitle{Computers \& Mathematics with Applications}
\bvolume{81},
\bfpage{336}--\blpage{354}
(\byear{2021})
\doiurl{10.1016/j.camwa.2020.12.019}
\end{barticle}
\endbibitem

\bibitem[\protect\citeauthoryear{Wang et~al.}{2025}]{pinn_review}
\begin{barticle}
\bauthor{\bsnm{Wang}, \binits{N.}},
\bauthor{\bsnm{Chen}, \binits{Y.}},
\bauthor{\bsnm{Zhang}, \binits{D.}}:
\batitle{A comprehensive review of physics-informed deep learning and its applications in geoenergy development}.
\bjtitle{The Innovation Energy}
\bvolume{2}(\bissue{2}),
\bfpage{100087}--\blpage{1}
(\byear{2025})
\end{barticle}
\endbibitem

\bibitem[\protect\citeauthoryear{Raissi et~al.}{2019}]{physicsinformeddeep}
\begin{barticle}
\bauthor{\bsnm{Raissi}, \binits{M.}},
\bauthor{\bsnm{Perdikaris}, \binits{P.}},
\bauthor{\bsnm{Karniadakis}, \binits{G.E.}}:
\batitle{Physics-informed neural networks: A deep learning framework for solving forward and inverse problems involving nonlinear partial differential equations}.
\bjtitle{Journal of Computational physics}
\bvolume{378},
\bfpage{686}--\blpage{707}
(\byear{2019})
\end{barticle}
\endbibitem

\bibitem[\protect\citeauthoryear{Lu et~al.}{2021}]{operatorlearning}
\begin{barticle}
\bauthor{\bsnm{Lu}, \binits{L.}},
\bauthor{\bsnm{Jin}, \binits{P.}},
\bauthor{\bsnm{Pang}, \binits{G.}},
\bauthor{\bsnm{Zhang}, \binits{Z.}},
\bauthor{\bsnm{Karniadakis}, \binits{G.E.}}:
\batitle{Learning nonlinear operators via deeponet based on the universal approximation theorem of operators}.
\bjtitle{Nature machine intelligence}
\bvolume{3}(\bissue{3}),
\bfpage{218}--\blpage{229}
(\byear{2021})
\end{barticle}
\endbibitem

\bibitem[\protect\citeauthoryear{Chen et~al.}{2018}]{ChenEtAl2018}
\begin{botherref}
\oauthor{\bsnm{Chen}, \binits{R.T.Q.}},
\oauthor{\bsnm{Rubanova}, \binits{Y.}},
\oauthor{\bsnm{Bettencourt}, \binits{J.}},
\oauthor{\bsnm{Duvenaud}, \binits{D.}}:
Neural Ordinary Differential Equations.
arXiv
(2018).
\doiurl{10.48550/ARXIV.1806.07366} .
\url{https://arxiv.org/abs/1806.07366}
\end{botherref}
\endbibitem

\bibitem[\protect\citeauthoryear{Ruthotto}{2024}]{ruthotto2024differential}
\begin{botherref}
\oauthor{\bsnm{Ruthotto}, \binits{L.}}:
Differential Equations for Continuous-Time Deep Learning
(2024)
\end{botherref}
\endbibitem

\bibitem[\protect\citeauthoryear{G{\"{u}}nther et~al.}{2021}]{gunther}
\begin{botherref}
\oauthor{\bsnm{G{\"{u}}nther}, \binits{S.}},
\oauthor{\bsnm{Pazner}, \binits{W.}},
\oauthor{\bsnm{Qi}, \binits{D.}}:
Spline parameterization of neural network controls for deep learning.
arXiv preprint arXiv:2103.00301
\textbf{abs/2103.00301}
(2021)
{\href{https://arxiv.org/abs/2103.00301}{{2103.00301}}}
\end{botherref}
\endbibitem

\bibitem[\protect\citeauthoryear{E}{2017}]{E:2017kz}
\begin{barticle}
\bauthor{\bsnm{E}, \binits{W.}}:
\batitle{{A Proposal on Machine Learning via Dynamical Systems}}.
\bjtitle{Communications in Mathematics and Statistics}
\bvolume{5}(\bissue{1}),
\bfpage{1}--\blpage{11}
(\byear{2017})
\doiurl{10.1007/s40304-017-0103-z}
\end{barticle}
\endbibitem

\bibitem[\protect\citeauthoryear{Haber and Ruthotto}{2017}]{HaberRuthotto2017}
\begin{barticle}
\bauthor{\bsnm{Haber}, \binits{E.}},
\bauthor{\bsnm{Ruthotto}, \binits{L.}}:
\batitle{{Stable architectures for deep neural networks}}.
\bjtitle{Inverse Problems}
\bvolume{34}(\bissue{1}),
\bfpage{014004}
(\byear{2017})
\doiurl{10.1088/1361-6420/aa9a90}
{\href{https://arxiv.org/abs/1705.03341}{{1705.03341}}}
\end{barticle}
\endbibitem

\bibitem[\protect\citeauthoryear{Ott et~al.}{2020}]{ott2021resnet}
\begin{botherref}
\oauthor{\bsnm{Ott}, \binits{K.}},
\oauthor{\bsnm{Katiyar}, \binits{P.}},
\oauthor{\bsnm{Hennig}, \binits{P.}},
\oauthor{\bsnm{Tiemann}, \binits{M.}}:
When are neural {ODE} solutions proper odes?
CoRR
\textbf{abs/2007.15386}
(2020)
{\href{https://arxiv.org/abs/2007.15386}{{2007.15386}}}
\end{botherref}
\endbibitem

\bibitem[\protect\citeauthoryear{Sander et~al.}{2022}]{sander2022residual}
\begin{botherref}
\oauthor{\bsnm{Sander}, \binits{M.E.}},
\oauthor{\bsnm{Ablin}, \binits{P.}},
\oauthor{\bsnm{Peyré}, \binits{G.}}:
Do Residual Neural Networks discretize Neural Ordinary Differential Equations?
(2022).
\url{https://arxiv.org/abs/2205.14612}
\end{botherref}
\endbibitem

\bibitem[\protect\citeauthoryear{Davis et~al.}{2020}]{autonomous}
\begin{botherref}
\oauthor{\bsnm{Davis}, \binits{J.Q.}},
\oauthor{\bsnm{Choromanski}, \binits{K.}},
\oauthor{\bsnm{Varley}, \binits{J.}},
\oauthor{\bsnm{Lee}, \binits{H.}},
\oauthor{\bsnm{Slotine}, \binits{J.E.}},
\oauthor{\bsnm{Likhosterov}, \binits{V.}},
\oauthor{\bsnm{Weller}, \binits{A.}},
\oauthor{\bsnm{Makadia}, \binits{A.}},
\oauthor{\bsnm{Sindhwani}, \binits{V.}}:
Time dependence in non-autonomous neural odes.
CoRR
\textbf{abs/2005.01906}
(2020)
{\href{https://arxiv.org/abs/2005.01906}{{2005.01906}}}
\end{botherref}
\endbibitem

\bibitem[\protect\citeauthoryear{Li et~al.}{2019}]{Li:2019wr}
\begin{botherref}
\oauthor{\bsnm{Li}, \binits{Q.}},
\oauthor{\bsnm{Lin}, \binits{T.}},
\oauthor{\bsnm{Shen}, \binits{Z.}}:
Deep learning via dynamical systems: An approximation perspective.
CoRR
\textbf{abs/1912.10382}
(2019)
{\href{https://arxiv.org/abs/1912.10382}{{1912.10382}}}
\end{botherref}
\endbibitem

\bibitem[\protect\citeauthoryear{Benning et~al.}{2019}]{benning2019deep}
\begin{barticle}
\bauthor{\bsnm{Benning}, \binits{M.}},
\bauthor{\bsnm{Celledoni}, \binits{E.}},
\bauthor{\bsnm{Ehrhardt}, \binits{M.J.}},
\bauthor{\bsnm{Owren}, \binits{B.}},
\bauthor{\bsnm{Sch{\"o}nlieb}, \binits{C.-B.}}:
\batitle{Deep learning as optimal control problems: Models and numerical methods}.
\bjtitle{Journal of Computational Dynamics}
\bvolume{6}(\bissue{2}),
\bfpage{171}--\blpage{198}
(\byear{2019})
\end{barticle}
\endbibitem

\bibitem[\protect\citeauthoryear{Newman et~al.}{2020}]{VarPro}
\begin{botherref}
\oauthor{\bsnm{Newman}, \binits{E.}},
\oauthor{\bsnm{Ruthotto}, \binits{L.}},
\oauthor{\bsnm{Hart}, \binits{J.L.}},
\oauthor{\bsnm{Bloemen~Waanders}, \binits{B.G.}}:
Train like a (var)pro: Efficient training of neural networks with variable projection.
CoRR
\textbf{abs/2007.13171}
(2020)
{\href{https://arxiv.org/abs/2007.13171}{{2007.13171}}}
\end{botherref}
\endbibitem

\bibitem[\protect\citeauthoryear{Zhong et~al.}{2019}]{zhong2019symplectic}
\begin{botherref}
\oauthor{\bsnm{Zhong}, \binits{Y.D.}},
\oauthor{\bsnm{Dey}, \binits{B.}},
\oauthor{\bsnm{Chakraborty}, \binits{A.}}:
Symplectic ode-net: Learning hamiltonian dynamics with control.
arXiv preprint arXiv:1909.12077
(2019)
\end{botherref}
\endbibitem

\bibitem[\protect\citeauthoryear{Nair et~al.}{2024}]{nair2024investigation}
\begin{bchapter}
\bauthor{\bsnm{Nair}, \binits{A.}},
\bauthor{\bsnm{Barwey}, \binits{S.}},
\bauthor{\bsnm{Pal}, \binits{P.}},
\bauthor{\bsnm{Maulik}, \binits{R.}}:
\bctitle{Investigation of latent time-scales in neural {ODE} surrogate models}.
In: \bbtitle{ICLR 2024 Workshop on AI4DifferentialEquations In Science}
(\byear{2024}).
\burl{https://openreview.net/forum?id=zLMeuYXUve}
\end{bchapter}
\endbibitem

\bibitem[\protect\citeauthoryear{Vermariën et~al.}{2025}]{Vermari_n_2025}
\begin{barticle}
\bauthor{\bsnm{Vermariën}, \binits{G.}},
\bauthor{\bsnm{Bisbas}, \binits{T.G.}},
\bauthor{\bsnm{Viti}, \binits{S.}},
\bauthor{\bsnm{Zhao}, \binits{Y.}},
\bauthor{\bsnm{Tang}, \binits{X.}},
\bauthor{\bsnm{Ravichandran}, \binits{R.}}:
\batitle{Neuralpdr: neural differential equations as surrogate models for photodissociation regions}.
\bjtitle{Machine Learning: Science and Technology}
\bvolume{6}(\bissue{2}),
\bfpage{025069}
(\byear{2025})
\doiurl{10.1088/2632-2153/ade4ee}
\end{barticle}
\endbibitem

\bibitem[\protect\citeauthoryear{Zhou and Barati~Farimani}{2025}]{Zhou_2025_neuralPDE}
\begin{barticle}
\bauthor{\bsnm{Zhou}, \binits{A.}},
\bauthor{\bsnm{Barati~Farimani}, \binits{A.}}:
\batitle{Predicting change, not states: An alternate framework for neural pde surrogates}.
\bjtitle{Computer Methods in Applied Mechanics and Engineering}
\bvolume{441},
\bfpage{117990}
(\byear{2025})
\doiurl{10.1016/j.cma.2025.117990}
\end{barticle}
\endbibitem

\bibitem[\protect\citeauthoryear{Yu et~al.}{2022}]{generalized}
\begin{botherref}
\oauthor{\bsnm{Yu}, \binits{D.}},
\oauthor{\bsnm{Miao}, \binits{H.}},
\oauthor{\bsnm{Wu}, \binits{H.}}:
Neural Generalized Ordinary Differential Equations with Layer-varying Parameters
(2022).
\url{https://arxiv.org/abs/2209.10633}
\end{botherref}
\endbibitem

\bibitem[\protect\citeauthoryear{Massaroli et~al.}{2020}]{dissect}
\begin{botherref}
\oauthor{\bsnm{Massaroli}, \binits{S.}},
\oauthor{\bsnm{Poli}, \binits{M.}},
\oauthor{\bsnm{Park}, \binits{J.}},
\oauthor{\bsnm{Yamashita}, \binits{A.}},
\oauthor{\bsnm{Asama}, \binits{H.}}:
Dissecting neural odes.
CoRR
\textbf{abs/2002.08071}
(2020)
{\href{https://arxiv.org/abs/2002.08071}{{2002.08071}}}
\end{botherref}
\endbibitem

\bibitem[\protect\citeauthoryear{He et~al.}{2015}]{heEtAl}
\begin{botherref}
\oauthor{\bsnm{He}, \binits{K.}},
\oauthor{\bsnm{Zhang}, \binits{X.}},
\oauthor{\bsnm{Ren}, \binits{S.}},
\oauthor{\bsnm{Sun}, \binits{J.}}:
Deep residual learning for image recognition.
CoRR
\textbf{abs/1512.03385}
(2015)
{\href{https://arxiv.org/abs/1512.03385}{{1512.03385}}}
\end{botherref}
\endbibitem

\bibitem[\protect\citeauthoryear{Huang et~al.}{2017}]{Huang_Liu_Lang_Yu_Wang_Li_2018}
\begin{botherref}
\oauthor{\bsnm{Huang}, \binits{L.}},
\oauthor{\bsnm{Liu}, \binits{X.}},
\oauthor{\bsnm{Lang}, \binits{B.}},
\oauthor{\bsnm{Yu}, \binits{A.W.}},
\oauthor{\bsnm{Li}, \binits{B.}}:
Orthogonal weight normalization: Solution to optimization over multiple dependent stiefel manifolds in deep neural networks.
CoRR
\textbf{abs/1709.06079}
(2017)
{\href{https://arxiv.org/abs/1709.06079}{{1709.06079}}}
\end{botherref}
\endbibitem

\bibitem[\protect\citeauthoryear{Vorontsov et~al.}{2017}]{orthogonalityrecurrentnetworks}
\begin{botherref}
\oauthor{\bsnm{Vorontsov}, \binits{E.}},
\oauthor{\bsnm{Trabelsi}, \binits{C.}},
\oauthor{\bsnm{Kadoury}, \binits{S.}},
\oauthor{\bsnm{Pal}, \binits{C.}}:
On orthogonality and learning recurrent networks with long term dependencies.
CoRR
\textbf{abs/1702.00071}
(2017)
{\href{https://arxiv.org/abs/1702.00071}{{1702.00071}}}
\end{botherref}
\endbibitem

\bibitem[\protect\citeauthoryear{Onken and Ruthotto}{2020}]{discoptvsoptdisc}
\begin{botherref}
\oauthor{\bsnm{Onken}, \binits{D.}},
\oauthor{\bsnm{Ruthotto}, \binits{L.}}:
Discretize-optimize vs. optimize-discretize for time-series regression and continuous normalizing flows.
CoRR
\textbf{abs/2005.13420}
(2020)
{\href{https://arxiv.org/abs/2005.13420}{{2005.13420}}}
\end{botherref}
\endbibitem

\bibitem[\protect\citeauthoryear{Kopp}{1962}]{Pontryagin_Max}
\begin{bchapter}
\bauthor{\bsnm{Kopp}, \binits{R.E.}}:
\bctitle{Pontryagin maximum principle}.
In: \beditor{\bsnm{Leitmann}, \binits{G.}} (ed.)
\bbtitle{Optimization Techniques}.
\bsertitle{Mathematics in Science and Engineering},
vol. \bseriesno{5},
pp. \bfpage{255}--\blpage{279}.
\bpublisher{Elsevier},
\blocation{New York}
(\byear{1962}).
\doiurl{10.1016/S0076-5392(08)62095-0} .
\burl{https://www.sciencedirect.com/science/article/pii/S0076539208620950}
\end{bchapter}
\endbibitem

\bibitem[\protect\citeauthoryear{Gholami et~al.}{2019}]{BIROS_ANODE}
\begin{botherref}
\oauthor{\bsnm{Gholami}, \binits{A.}},
\oauthor{\bsnm{Keutzer}, \binits{K.}},
\oauthor{\bsnm{Biros}, \binits{G.}}:
{ANODE:} unconditionally accurate memory-efficient gradients for neural odes.
CoRR
\textbf{abs/1902.10298}
(2019)
{\href{https://arxiv.org/abs/1902.10298}{{1902.10298}}}
\end{botherref}
\endbibitem

\bibitem[\protect\citeauthoryear{Abraham et~al.}{1993}]{abraham_manifolds_1993}
\begin{bbook}
\bauthor{\bsnm{Abraham}, \binits{R.}},
\bauthor{\bsnm{Marsden}, \binits{J.E.}},
\bauthor{\bsnm{Ratiu}, \binits{T.}}:
\bbtitle{Manifolds, {Tensor} {Analysis}, and {Applications}}.
\bsertitle{Applied {Mathematical} {Sciences}}.
\bpublisher{Springer},
\blocation{New York}
(\byear{1993}).
\burl{https://books.google.com/books?id=dWHet\_zgyCAC}
\end{bbook}
\endbibitem

\bibitem[\protect\citeauthoryear{Calcaterra and Boldt}{2006}]{calcaterra2006lipschitzflowboxtheorem}
\begin{botherref}
\oauthor{\bsnm{Calcaterra}, \binits{C.}},
\oauthor{\bsnm{Boldt}, \binits{A.}}:
Lipschitz Flow-box Theorem
(2006).
\url{https://arxiv.org/abs/math/0305207}
\end{botherref}
\endbibitem

\bibitem[\protect\citeauthoryear{Khalil}{2002}]{Khalil_2002}
\begin{bbook}
\bauthor{\bsnm{Khalil}, \binits{H.K.}}:
\bbtitle{Lyapunov Stability},
\bedition{3}rd edn.,
pp. \bfpage{112}--\blpage{140}.
\bpublisher{Prentice Hall}, \blocation{???}
(\byear{2002})
\end{bbook}
\endbibitem

\bibitem[\protect\citeauthoryear{Ruthotto}{2020}]{ruthotto2020numerical}
\begin{botherref}
\oauthor{\bsnm{Ruthotto}, \binits{L.}}:
A Numerical Analysis Perspective on Deep Neural Networks.
YouTube, NeurIPS 2020 Workshop on Differentiable Programming
(2020).
\url{https://www.youtube.com/watch?v=xL2KZZMrPwA}
\end{botherref}
\endbibitem

\bibitem[\protect\citeauthoryear{Kingma and Ba}{2014}]{adam}
\begin{botherref}
\oauthor{\bsnm{Kingma}, \binits{D.P.}},
\oauthor{\bsnm{Ba}, \binits{J.}}:
Adam: A method for stochastic optimization.
arXiv preprint arXiv:1412.6980
(2014)
{\href{https://arxiv.org/abs/1412.6980}{{arXiv:1412.6980}}}
{[cs.LG]}
\end{botherref}
\endbibitem

\bibitem[\protect\citeauthoryear{He et~al.}{2019}]{zhouNODES}
\begin{bchapter}
\bauthor{\bsnm{He}, \binits{L.}},
\bauthor{\bsnm{Xie}, \binits{X.}},
\bauthor{\bsnm{Lin}, \binits{Z.}}:
\bctitle{Neural ordinary differential equations with envolutionary weights}.
In: \bbtitle{Pattern Recognition and Computer Vision: Second Chinese Conference, PRCV 2019, Xi'an, China, November 8-11, 2019, Proceedings, Part I},
pp. \bfpage{598}--\blpage{610}.
\bpublisher{Springer},
\blocation{Berlin, Heidelberg}
(\byear{2019}).
\doiurl{10.1007/978-3-030-31654-9\_51} .
\burl{https://doi.org/10.1007/978-3-030-31654-9\_51}
\end{bchapter}
\endbibitem

\end{thebibliography}

\end{document}